\documentclass[letterpaper, 10 pt, conference]{ieeeconf}  

\IEEEoverridecommandlockouts                              
\usepackage[linesnumbered,ruled]{algorithm2e}
\usepackage{float}
\usepackage{subfigure}
\usepackage{graphics} 
\usepackage{epsfig} 
\usepackage{mathrsfs}
\usepackage{times} 
\usepackage{amsmath} 
\usepackage{amssymb}  
\usepackage{mathrsfs}
\usepackage{color}
\usepackage[table]{xcolor}
\usepackage[T1]{fontenc}
\usepackage{bm}
\usepackage{cite}
\usepackage{hyperref}
\usepackage{booktabs}
\usepackage{threeparttable}
\usepackage{multirow}
\hypersetup{
	colorlinks=true,
	linkcolor=blue,
	filecolor=blue,      
	urlcolor=blue,
	citecolor=cyan,
}

\title{\LARGE \bf
    Automatic Failure Recovery and Re-Initialization for Online UAV Tracking with Joint Scale and Aspect Ratio Optimization
	}
\author{Fangqiang Ding$^{1}$, Changhong Fu$^{1,*}$, Yiming Li$^{1}$, Jin Jin$^{1}$ and Chen Feng$^{2}$ 
	\thanks{*Corresponding author}
	\thanks{$^{1}$Fangqiang Ding, Changhong Fu, Yiming Li and Jin Jin are with the School of Mechanical Engineering, Tongji University, 201804 Shanghai, China.
		{\tt\small changhongfu@tongji.edu.cn}}
		\thanks{$^{2}$Chen Feng is with the Tandon School of Engineering, New York University, NY 11201 New York, United States.
		{\tt\small cfeng@nyu.edu}
	}
\thanks{The code and tracking video are respectively released on \url{http://github.com/vision4robotics/JSAR-Tracker} and \url{https://youtu.be/wNFhkpH6osk}.}
}

\begin{document}
\maketitle
\thispagestyle{empty}
\pagestyle{empty}

\begin{abstract}
Current unmanned aerial vehicle (UAV) visual tracking algorithms are primarily limited with respect to: (i) the kind of size variation they can deal with, (ii) the implementation speed which hardly meets the real-time requirement. In this work, a real-time UAV tracking algorithm with powerful size estimation ability is proposed. Specifically, the overall tracking task is allocated to two 2D filters: (i) translation filter for location prediction in the space domain, (ii) size filter for scale and aspect ratio optimization in the size domain. Besides, an efficient two-stage re-detection strategy is introduced for long-term UAV tracking tasks. Large-scale experiments on four UAV benchmarks demonstrate the superiority of the presented method which has computation feasibility on a low-cost CPU.
\end{abstract}
\section{INTRODUCTION}\label{sec:INTRODUCTION}
Equipped with visual perception capability, robots can have flourishing real-world applications, \emph{e.g.}, visual object tracking has stimulated broad practical utilities like human-robot collaboration~\cite{palinko2016robot}, robotic arm manipulation~\cite{hofer2019iterative}, and aerial filming~\cite{bonatti2019towards}.
Tracking onboard unmanned aerial vehicle (UAV) has many advantages over general object tracking, for instance, broad view scope, high flexibility, and mobility. Yet more difficulties are introduced such as aspect ratio change (ARC)\footnote{Caused by rapid attitude alteration and strong motion of UAV, ARC is generally brought with the form of large viewpoint variation, intense rotation, deformation, to name a few.}, out-of-view, exiguous calculation resources, \emph{etc}. Hence, a robust-against-ARC, low-cost, and energy-efficient tracking algorithm applicable in long short-term tasks is highly desirable for UAV tracking.

In literature, although deep feature~\cite{danelljan2016beyond,li2017integrating, danelljan2017eco} or deep architecture~\cite{bertinetto2016fully,li2019target,guo2017learning} can exceedingly boost the tracking robustness, the complex convolution operations have hampered their practical utility. Another research direction in visual tracking is discriminative correlation filters (DCF)~\cite{henriques2015high,bertinetto2016staple,galoogahi2017learning,Li2020CVPR}. With only hand-crafted features, DCF-based trackers mostly have real-time speed on a single CPU thanks to their transforming intractable spatial convolution into element-wise multiplication in the Fourier domain.
\begin{figure}[!t]
	\includegraphics[width=0.98\columnwidth]{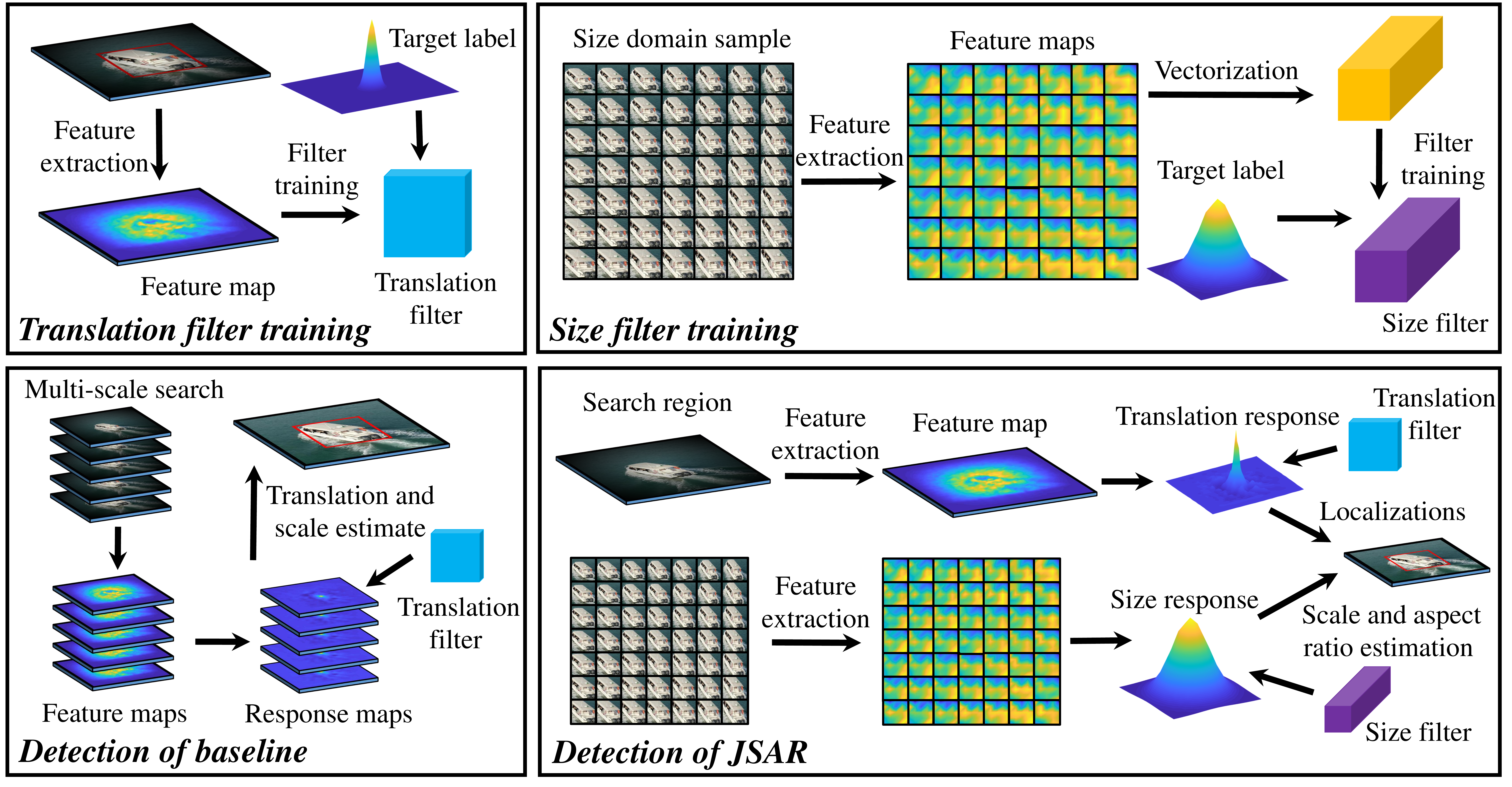}
	\caption{Comparison of overall flow in baseline \cite{li2018CVPR} (left) and JSAR (right). Baseline only trains a translation filter for translation estimation in the detection phase and updates object scale by brute-force multi-scale search strategy. JSAR proposes to train size filter in size domain by multi-size sampling and jointly estimate the scale and aspect ratio of the object in detection stage.}
	\label{fig:1}
\end{figure}
While most researches focus on location and scale estimation, scarce of them focus on aspect ratio. Current DCF-based trackers commonly fix the aspect ratio of the object during tracking. Consequently, in UAV tracking scenarios with extensive ARC, erroneous appearance is frequently introduced in filter training because of the imprecise size estimation, leading to filter degradation. 

Inspired by 1D scale filter~\cite{danelljan2017discriminative} aiming to handle inefficiency of brute-force multi-scale search~\cite{li2014scale,danelljan2015learning,galoogahi2017learning,Li2020TSD}, this work proposes a \textbf{j}oint \textbf{s}cale and \textbf{a}spect \textbf{r}atio optimization tracker (JSAR) to achieve accurate scale and aspect ratio estimation. As displayed in Figure~\ref{fig:1}, the training procedure is two-fold: (i) training a translation filter with a single patch \cite{li2018CVPR} for location prediction, and (ii) training a size filter with exponentially-distributed samples for scale and aspect ratio estimation. Consequently, location, scale, and aspect ratio are calculated simultaneously, \emph{i.e.}, the object bounding box can be estimated in the 4-DoF (degree of freedom) space, promoting the tracking accuracy without losing much speed.

Recently, combining visual tracking with re-detection framework has raised precision in the long-term tracking scenarios where objects frequently suffer from out-of-view or full occlusion~\cite{fan2017parallel,ma2015long}. Yet the speed is mostly sacrificed due to the intractable object detection methods. In this work, a CPU-friendly re-detection strategy is proposed to enable long-term tracking. An effective tracking failure monitoring mechanism and an efficient re-initialization method based on EdgeBoxes~\cite{zitnick2014edge} collaboratively contribute to the smooth long-term tracking. Our main contributions are three-fold:
\begin{itemize}
	\item A novel robust tracking method with real-time speed is proposed with joint scale and aspect ratio optimization.
	\item A new CPU-friendly re-detection framework is developed to accomplish long-term tracking tasks efficiently.
	\item Large-scale experiments conducted on three short-term UAV benchmarks and one long-term benchmark validate the outstanding performance of the proposed method. 
\end{itemize}
\section{RELATED WORKS}\label{sec:RELATEDWORK}
\subsection{Discriminative correlation filter tracking algorithm}
In literature, discriminative tracking algorithms train a classifier to differentiate the tracked object from the background by maximizing the classification score. Recent investigations focus on discriminative correlation filters since D. S. Bolme \emph{et al.}~\cite{bolme2010visual} proposed to learn robust filters by mapping the training samples to the desired output. J. F. Henriques \emph{et al.}~\cite{henriques2015high} presented to solve the rigid regression equation in the Fourier domain, and established the basic structure of modern DCF methods. Afterwards, several attempts are made to promote tracking performance within DCF framework, \emph{e.g.}, spatial penalization~\cite{danelljan2015learning, dai2019visual}, multi-feature fusion~\cite{wang2018multi,bertinetto2016staple}, and real negative sampling~\cite{galoogahi2017learning,mueller2017context}. However, most research highlight the improvement of localization accuracy rather than amelioration in the size estimation.
\subsection{Prior works in object size estimation}
Pioneer DCF trackers\cite{bolme2010visual,henriques2012exploiting,henriques2015high} fix the object size and only estimate the trajectory in the 2D space. Presetting a scaling pool, \cite{li2014scale} sampled on different scales to find the optimal one in the detection phase. \cite{danelljan2014accurate} proposed a separate scale correlation filter to estimate scale variance in the 1D scale domain. To enable aspect ratio estimation, \cite{huang2015enable} tackled scale and aspect ratio variation by embedding detection proposals generator in tracking pipeline. \cite{li2017integrating} enforced near-orthogonality constraint on center and boundary filters. Despite bringing more freedoms in object tracking, these two methods bring heavy computation burden for DCF trackers, and are hence not satisfactory alternatives for UAV tracking.  
\subsection{Re-detection in object tracking}
Tracking-learning-detection (TLD) \cite{kalal2012tracking} is proposed to validate tracking results and decide whether to enable learning and detection. Among DCF trackers,  \cite{ma2015long} introduced an online random fern to generate candidates and score each of them for re-detection. Despite the effectiveness, it is time-consuming due to the scanning window strategy. \cite{fan2017parallel} presented a novel multi-threading framework in which an offline-trained Siamese network is used as a verifier. However, speed is largely decreased. This work utilizes EdgeBoxes~\cite{zitnick2014edge} to quickly generate proposals, and then a decision filter is applied to select the most possible bounding box for tracker's re-initialization. The proposed two-stage re-detection strategy is more efficacious and light-weight. 
\subsection{UAV tracking}
In UAV tracking scenarios, the tracked objects possess higher motion flexibility than in tracking based on hand-held or fixed surveillance cameras. Therefore, UAV tracking is confronted with more difficulties. In literature, aberrance repression \cite{huang2019learning} and intermittent context learning \cite{li2020keyfilter, li2020intermittent} are proposed to improve tracking precision. Despite obtaining appealing results, they cannot estimate aspect ratio variation. Adaptive to ARC, JSAR has better robustness and real-time speed, and hence superior to other trackers in UAV tracking.

\section{PROPOSED METHOD} 
\subsection{Discriminative correlation filter}
In frame $t$, a multi-channel correlation filter $\mathbf{W}_t\in\mathbb{R}^{M\times N\times D}$ is trained by restricting its correlation result with training samples $\mathbf{X}_t\in\mathbb{R}^{M\times N\times D}$ to the given target label $\mathbf{g}\in\mathbb{R}^{M\times N}$. The minimized objective $\mathcal{E}(\mathbf{W}_t)$ is formulated as the sum of least square term and  regularization term:
\begin{equation}\small\label{eqn:1}
\mathcal{E}(\mathbf{W}_t)=\Big\Vert\sum_{d=1}^{D}\mathbf{w}_t^d\star\mathbf{x}_t^d-\mathbf{g}\Big\Vert^2_2+\lambda\sum_{d=1}^{D}\Big\Vert\mathbf{w}_t^d\Big\Vert^2_2~,
\end{equation}
\noindent where $\mathbf{x}_t^d\in\mathbb{R}^{M\times N}$ and $\mathbf{w}_t^d\in\mathbb{R}^{M\times N}$ respectively indicate the \emph{d}-th channel feature representation and filter, and $\star$ denotes the cyclic correlation operator. $M$ and $N$ denote the width and height of a single channel sample while $D$ denotes the number of feature channels. $\lambda$ is a hyper parameter for avoiding over-fitting. Minimizing the objective in the Fourier domain, a closed-form solution of filter $\mathbf{W}_t$ is obtained:
\begin{equation}\small\label{eqn:2}
	\widetilde{\mathbf{w}}^d_t=\frac{\widetilde{\mathbf{g}}\odot\widetilde{\mathbf{x}}^{d*}_{t}}{\sum_{d=1}^{D}(\widetilde{\mathbf{x}^d_t}\odot\widetilde{\mathbf{x}}^{d*}_{t})+\lambda}~,
\end{equation}
\noindent where $\odot$ and $\frac{~\cdot~}{~\cdot~}$ denote the element-wise multiplication and division, respectively. $\widetilde{\cdot}$ means discrete Fourier transform (DFT) and $\cdot^{*}$ means complex conjugation. The appearance model is updated by linear interpolation with a predefined learning rate $\theta$. Use $\mathcal{F}^{-1}$ to denote inverse discrete Fourier transform and $\mathbf{m}^d_t$ as the \emph{d}-th feature representation of the search region, the response map $\mathbf{R}_t$ is obtained by:
\begin{equation}\small\label{eqn:4}
\mathbf{R}_t=\mathcal{F}^{-1}\Big(\sum_{d=1}^{D}
\widetilde{\mathbf{w}}_{t-1}^{d*}\odot\widetilde{\mathbf{m}}^d_t\Big)~.	
\end{equation} 
\subsection{Translation estimation}
For translation estimation, most trackers~\cite{henriques2015high,wang2018multi,bertinetto2016staple} learn a 2D translation filter $\mathbf{W}_{t,trans}$ in the space domain by Eq.~(\ref{eqn:2}). In the training phase, the region of interest (ROI) is cropped centered at object location with a fixed proportion to the object scale. In the detection phase, the feature of ROI centered at the location of the last frame $\mathbf{M}_{t,trans}$ is extracted. The object is localized by finding the peak position of the response map generated by Eq.~(\ref{eqn:4}). For scale estimation, classical brute-force search in a multi-scale hierarchical structure is inefficient due to repetitive feature extraction on large image patches.
\subsection{Size estimation}
Motivated by~\cite{danelljan2014accurate} which trains a 1D scale filter in the scale domain for scale-adaptive tracking, we propose to train a 2D size filter $\mathbf{W}_{t,size}$. Different to the 2D translation filter learned in the space domain which is composed of a horizontal and vertical axis, the samples are extracted in the size domain consisting of a scale axis and aspect ratio axis. 
 \begin{figure*}[!t]
	\centering
	\includegraphics[width=\textwidth]{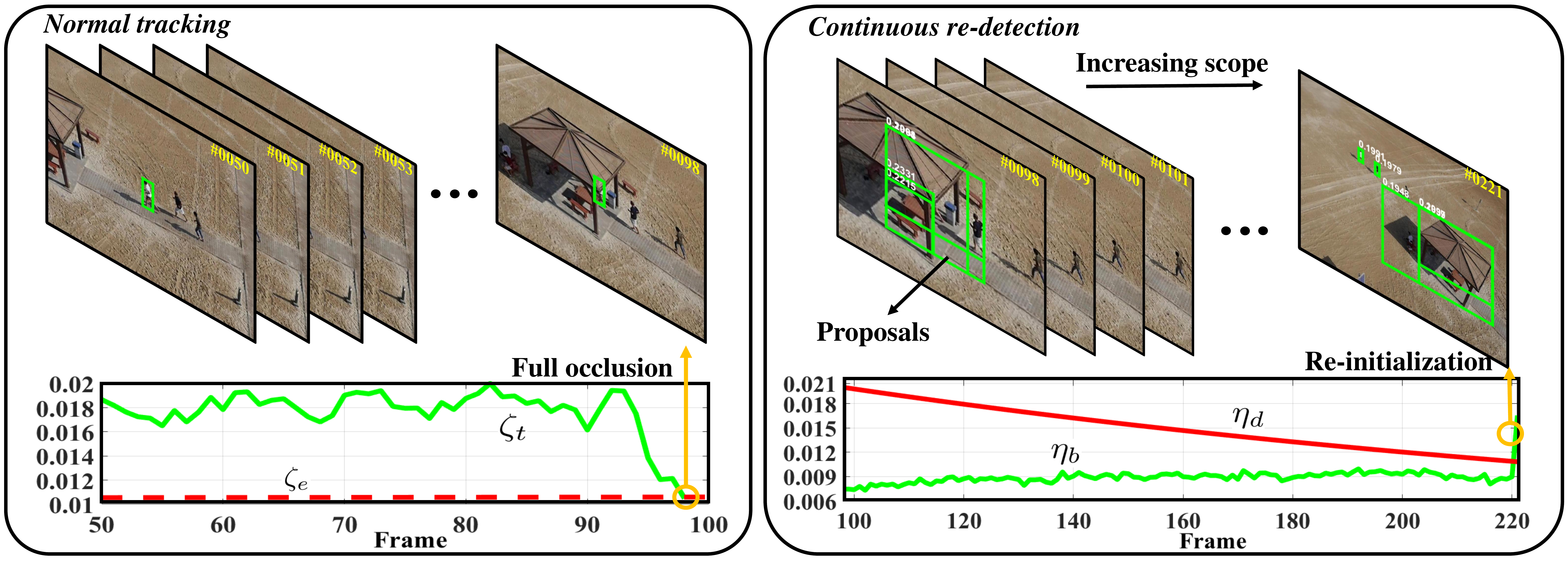}
	\caption{Overall flow of re-detection strategy. When the maximum value of the response map in frame $t$ ($\zeta_t$)  is larger than the threshold value ($\zeta_e$), tracking procedure is normally implemented, if not, re-detection mechanism is activated. When the peak value of the response map $\eta_b$ generated by selected proposal and decision filter exceeds the descending threshold $\eta_d$, the bounding box is re-initialized and re-starts to be tracked normally. It is note that the values remarked above proposals are confidence scores $k^i$ and the search scope is increasing in continuous re-detection. }
	\label{fig:2}
\end{figure*}
\subsubsection{Sampling in size domain}
During size filter training, we crop $S\times A$ patches centered at current object location, where $A$ and $S$ represent the number of aspect ratios and scales in training sample. The size of these patches is calculated by:
\begin{equation}\small\label{eqn:5}
\begin{aligned}
\{W^{s,a}_t,H^{s,a}_t\}=\{W_t\gamma^{N_s}\phi^{N_a},H_t\gamma^{N_s}/\phi^{N_a}\}\\
(s=1,2,\cdots,S,~a=1,2,\cdots,A)
\end{aligned}
~,
\end{equation} 
\noindent where $W_t$ and $H_t$ are the object width and height in frame $t$, and \{$s,a$\} denotes the index of patches with various scale and aspect ratio. To maintain sample symmetry, we set $N^s=-\frac{S+1}{2}+s$ and $N^a=-\frac{A+1}{2}+a$. Otherwise, $\gamma$ and $\phi$ is hyper parameters to control sampling step. To make sure the dimension consistency of cropped patches and reduce computation burden, all patches are downsampled to a presetting model size \{$W_{model},H_{model}$\}. Afterwards, feature map $\mathbf{V}_t^{s,a}\in\mathbb{R}^{\frac{W_{model}}{C}\times \frac{H_{model}}{C}\times K}$ is extracted on each patch with $K$ feature channels. Here, $C$ denotes the side length of single cell for feature extraction. Different to translation filter which utilizes original feature map for training, for each patch, the extracted feature representations are vectorized to 1D vector $\mathbf{v}^{s,a}_t=\textbf{vec}(\mathbf{V}_t^{s,a})\in\mathbb{R}^{\frac{W_{model}H_{model }K}{C^2}}$, as exhibited in Fig. \ref{fig:1}. In this process, the number of feature channel changes from original $K$ to $C=\frac{W_{model}H_{model }K}{C^2}$. By stacking column vectors from different patches, the final sample $\mathbf{X}_{t,size}\in\mathbb{R}^{S\times A\times C}$ can be denoted by:
\begin{equation}\footnotesize\label{eqn:6}
\mathbf{X}_{t,size}=
\begin{bmatrix}
\mathbf{v}^{1,1}_t&\mathbf{v}^{1,2}_t&\mathbf{v}^{1,3}_t&\cdots&\mathbf{v}^{1,A}_t\\
\mathbf{v}^{2,1}_t&\mathbf{v}^{2,2}_t&\mathbf{v}^{2,3}_t&\cdots&\mathbf{v}^{2,A}_t\\
\mathbf{v}^{3,1}_t&\mathbf{v}^{3,2}_t&\mathbf{v}^{3,3}_t&\cdots&\mathbf{v}^{3,A}_t\\
\vdots&\vdots&\vdots&\ddots&\vdots\\
\mathbf{v}^{S,1}_t&\mathbf{v}^{S,2}_t&\mathbf{v}^{S,3}_t&\cdots&\mathbf{v}^{S,A}_t\\
\end{bmatrix}~.
\end{equation}
\subsubsection{Size estimation}
After sample extraction, Eq. (\ref{eqn:2}) is applied to learn size filter $\mathbf{W}_{t,size}$. In the estimation stage, we assume the size is unchanged and estimate the location translation at first when a new frame comes. Centering at the predicted location, the feature representation of the search region in size domain $\mathbf{M}_{t,size}$ is extracted for size estimation, as shown in Fig. \ref{fig:1}. It is noted that $\mathbf{M}_{t,size}$ has the same dimension as training sample $\mathbf{X}_{t,size}$. By Eq.~(\ref{eqn:4}), the current scale and aspect ratio are obtained by maximizing the response score, and then the object size is optimized. In a word, our method can be generally applicable in the DCF framework, and work in the 4-DoF space.
\subsection{Re-detection strategy}
As displayed in Fig.~\ref{fig:2}, the re-detection will be implemented when tracking failure is observed. The proposed re-detection strategy has two stages, \emph{i.e.}, object proposals generation, and candidates scoring. Object proposal method EdgeBoxes [22] is applied to generate candidates and decision filter scores them for object re-initialization. The detailed illustration is as follows.
\subsubsection{Tracking failure monitoring mechanism}
Ideally, the re-detection is enabled when the object is lost or the output deviates greatly against real object location. Related to the tracking confidence, the peak value $\zeta_t=max(\mathbf{R}_{t,trans})$ of response map generated in translation estimation of frame $t$ is adopted to decide whether to activate the re-detection. Presetting a threshold $\zeta_e$, the re-detection mechanism is activated when $\zeta_t<\zeta_e$. 
\subsubsection{Object proposal generation}
When re-detection begins, EdgeBoxes~\cite{zitnick2014edge} is utilized to generate class-agnostic object proposals within surrounding square area at the first stage. The side length of this surrounding area is $\omega\sqrt{W_tH_t}$ in this work. Each proposal $\mathbf{b}^i$ generated by EdgeBoxes has five variables, \emph{i.e.}, $\mathbf{b}^i=[x^i,y^i,w^i,h^i,k^i]$. The first four variables denote the location and size of the proposal while the last value $k_i$ is the confidence score. Depending on the confidence score, we choose the top $N_e$ proposals for re-detection.
\subsubsection{Object proposals scoring}
During the second stage, for each proposal, the feature from ROI in frame $t$ is extracted with $K$ feature channels, which is denoted by $\mathbf{P}_t^i   (i=1,2,\cdots, N_e)$. To make a final decision for re-initialization, a decision filter $\mathbf{M}_{deci}$ is trained along with translation filter using selected pure samples\footnote{ The selection of pure samples also depends on the peak value of response map in frame $t$: if the peak value $\zeta_t>\zeta_s$ ($\zeta_s$ is a predefined threshold), the sample for translation filter training is adopted to update the decision filter because larger peak value indicates better tracking quality.}. After feature extraction of proposals $\mathbf{P}_t^i   (i=1,2,\cdots, N_e)$, the corresponding $N_e$ response maps are calculated through the correlation of decision filter and feature maps by Eq.~(\ref{eqn:4}), and then the proposal with the largest peak value is selected. However, in the scenarios of out-of-view and full occlusion, the selected proposal is generally fallacious. To this end, we set a threshold $\eta_d$ to decide whether to re-initialize: if the selected proposal's peak value $\eta_b>\eta_d$, the re-initialization will be enabled, or else, re-detection is continued. 

In this work, the scale of the search area is increased and the re-initialization threshold $\eta_d$ is reduced frame-by-frame in re-detection failure cases to make sure the re-initialization ultimately works. The overall flow of the proposed method is presented in Algorithm~\ref{alg:KAOTtrackerflow}.
\begin{algorithm}[!t]
	\label{alg:KAOTtrackerflow}
	\caption{JSAR-Re}
	\KwIn{Object location and size at the first frame\\
		\hspace{1.1cm}Subsequent images in the video sequence
	}	
	\KwOut{Location and size of object in frame $t$}
	\label{alg:workflow}
	\eIf{$t=1$ or re-detection enabled}{
		Extract training samples $\mathbf{X}_{i,trans}$ and $\mathbf{X}_{i,size}$\\
		Use Eq.~(\ref{eqn:2}) to initialize $\mathbf{W}_{i,trans}$ and $\mathbf{W}_{i,size}$\\
		Initialize $\mathbf{M}_{deci}$ by $\mathbf{X}_{i,trans}$,  disable re-detection
	}
	{  
		Extract search region feature maps $\mathbf{M}_{t,trans}$\\
		Generate $\mathbf{R}_{t,trans}$ by  Eq.~(\ref{eqn:4}) and find $\zeta_t$\\
		\eIf{$\zeta_t>\zeta_e$}{
			Estimate object translation and extract $\mathbf{M}_{i,size}$\\
			Estimate object size using Eq.~(\ref{eqn:4})\\
			Use Eq.~(\ref{eqn:2}) to update $\mathbf{W}_{t,trans}$ and $\mathbf{W}_{t,size}$\\
			\If{$\zeta_t>\zeta_s$}{
				Update filter $\mathbf{M}_{deci}$ using $\mathbf{M}_{t,trans}$}
		}
		{
			Generate proposals with search area\\
			Scoring proposals using $\mathbf{M}_{deci}$ by Eq.~(\ref{eqn:4})\\
			Find the largest peak value $\eta_b$
			\\
			\eIf{$\eta_b>\eta_d$}
			{Enable re-detection, initialize the object}
			{Increase $\omega$ and reduce $\eta_d$\\
				Continue to re-detect next frame	
			}
		}
	}	
\end{algorithm}
\begin{table}[!b]
	\centering
	\setlength{\tabcolsep}{1.8mm}
	\fontsize{8}{9}\selectfont
	\begin{threeparttable}
		\caption{For impartial comparison, these parameters are fixed in all evaluation of our trackers}
		\vspace{0.08cm}
		\begin{tabular}{ccc}
			\toprule[2pt]
			Symbol&Value&Meaning\\
			\midrule
			S&13&The number of scales\\
			A&13&The number of aspect ratios\\
			$\gamma$&1.03&Scale sampling step\\
			$\phi$&1.02&Aspect ratio sampling step\\
			$\theta_{size}$&0.014&The learning rate of size filter\\
			$W_{model}$&16&The width of model size\\
			$H_{model}$&32&The height of mode size\\
			$C$&4&The side length of feature cell\\
			$\zeta_e$&0.0105&Re-detection enablement threshold\\
			$\zeta_s$&0.013&Decision filter update threshold\\
			$\eta_d$&0.02&Re-initialization threshold\\
			$\omega$&5&The side length factor of re-detection area\\
			$N_e$&30&The number of proposals for re-detection\\
			\bottomrule[2pt]
			\label{tab:4}
		\end{tabular}
	\end{threeparttable}
\end{table}
\section{EXPERIMENTS}
\begin{figure*}[!t]
	\begin{center}
		\subfigure { \label{fig:UAVDT} 
			\begin{minipage}{0.315\textwidth}
				\centering
				\includegraphics[width=1\columnwidth]{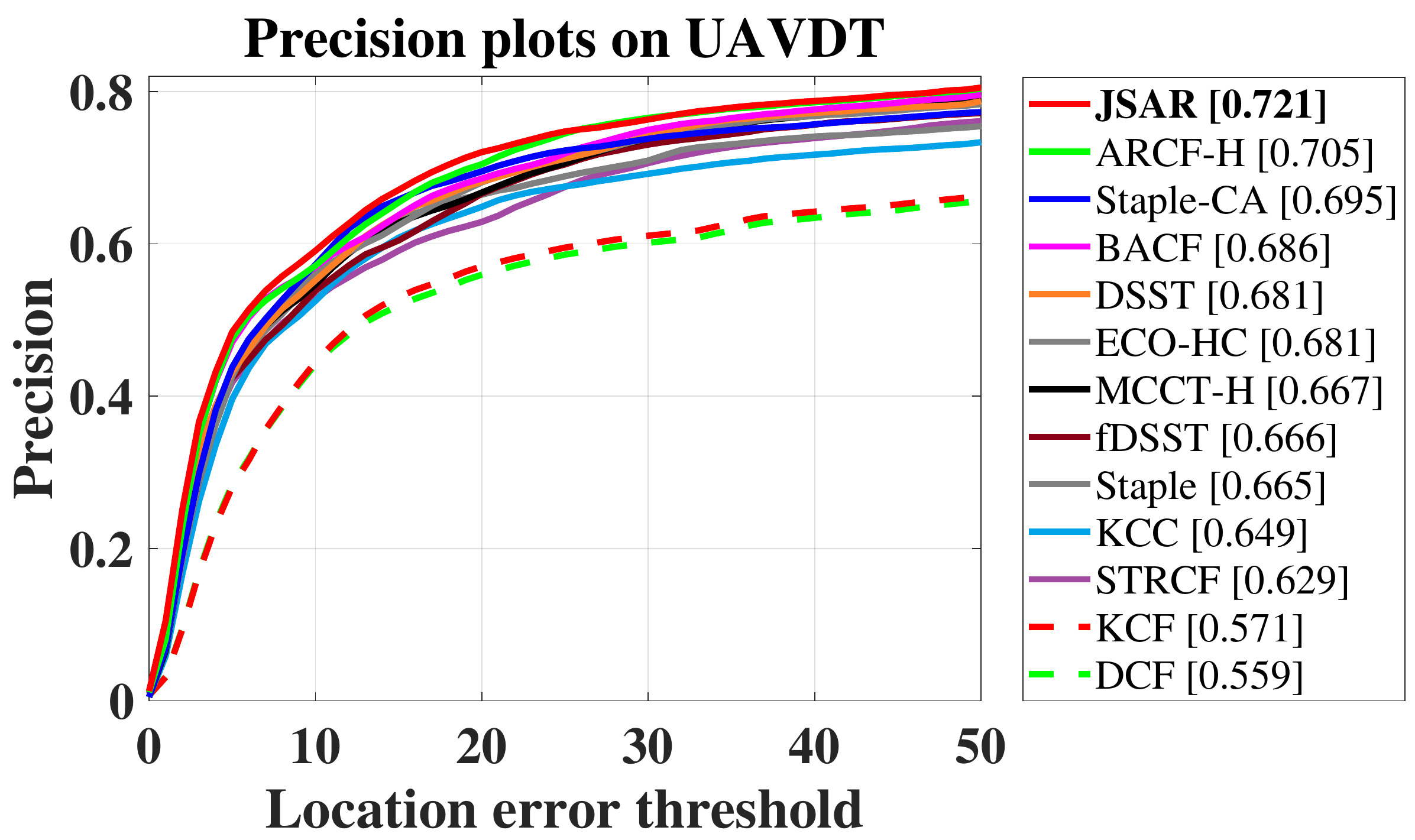}
				\\
				\includegraphics[width=1\columnwidth]{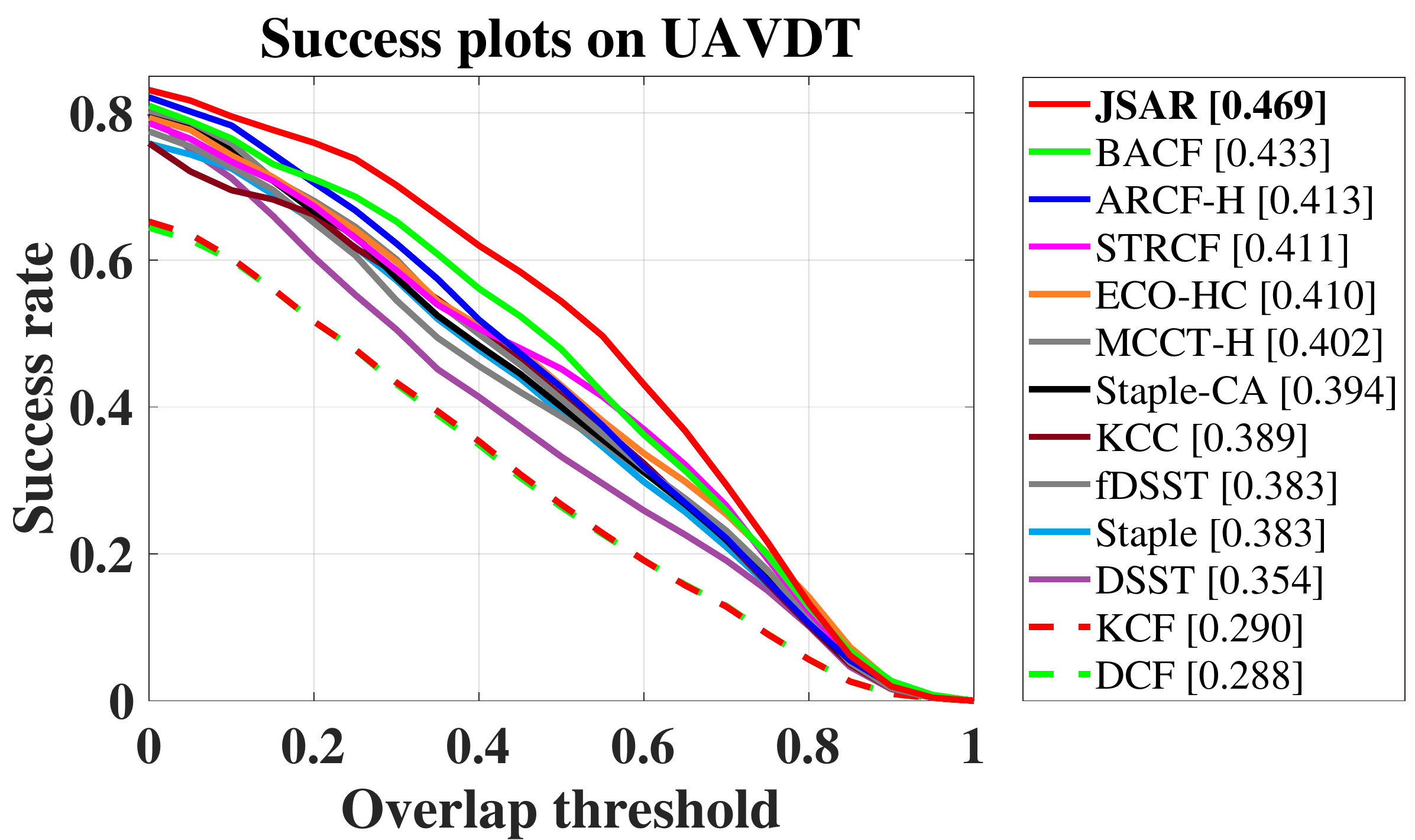}
			\end{minipage}
		}
		\subfigure { \label{fig:UAV123} 
			\begin{minipage}{0.315\textwidth}
				\centering
				\includegraphics[width=1\columnwidth]{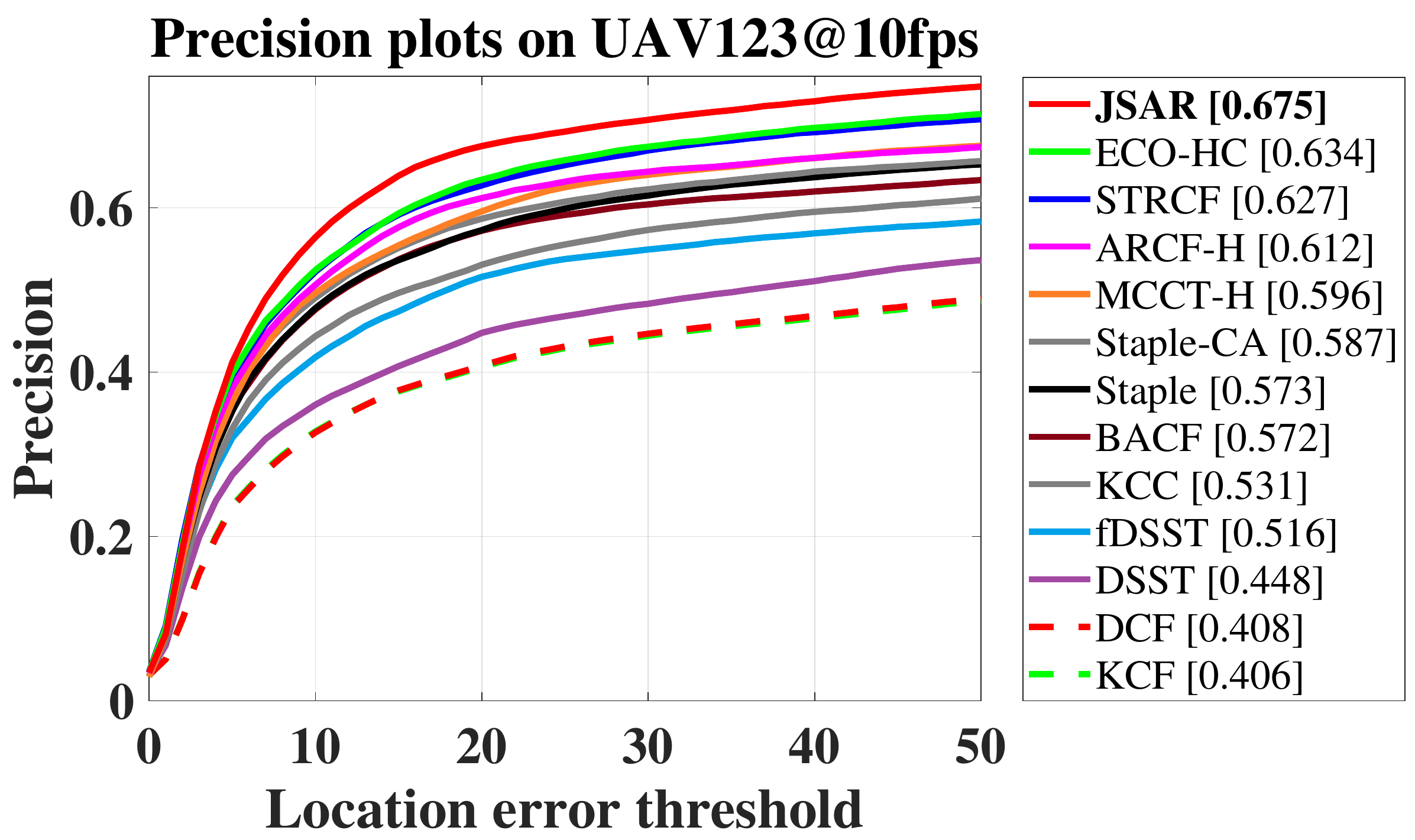}
				\\
				\includegraphics[width=1\columnwidth]{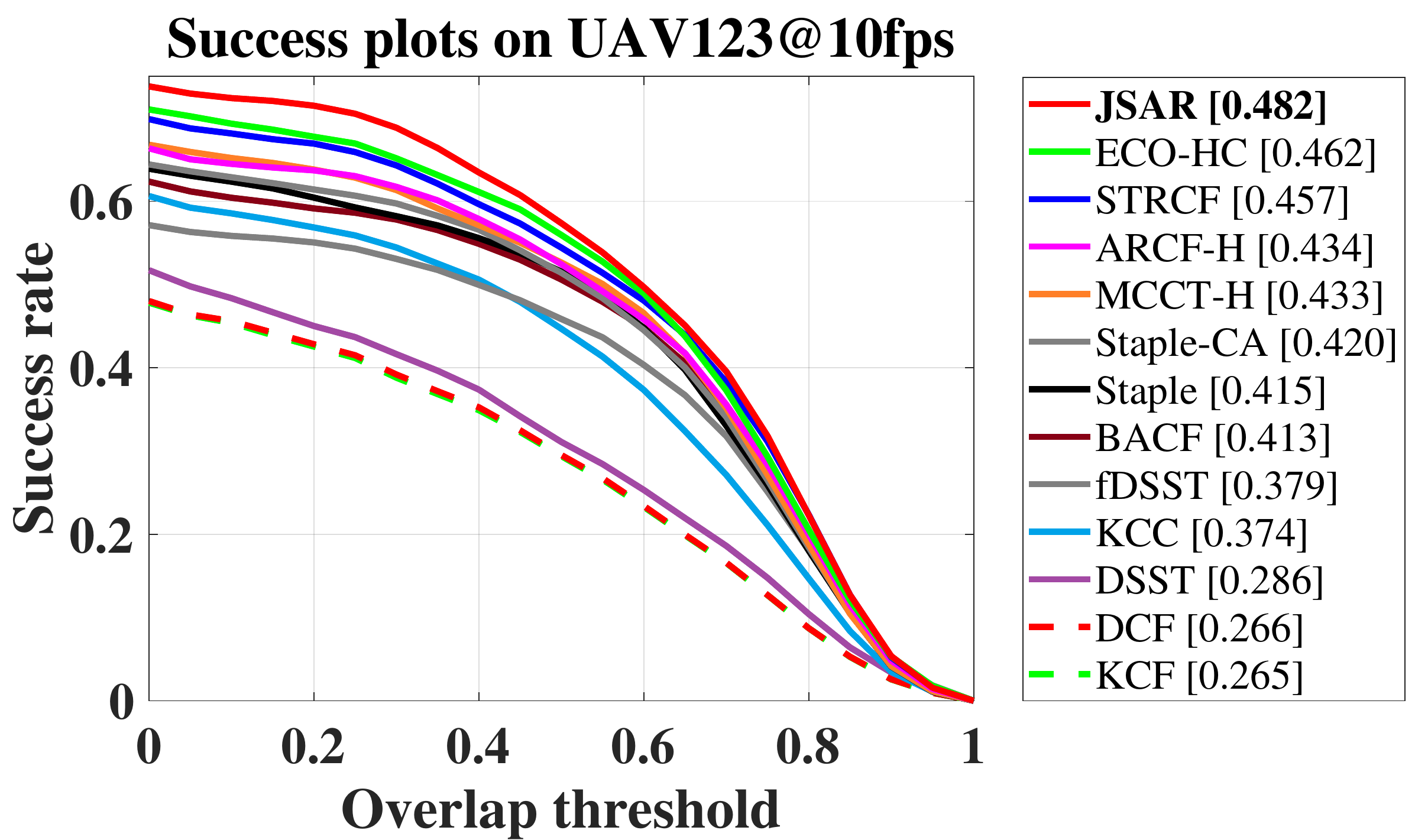}
			\end{minipage}
		}
		\subfigure { \label{fig:DTB70} 
			\begin{minipage}{0.315\textwidth}
				\centering
				\includegraphics[width=1\columnwidth]{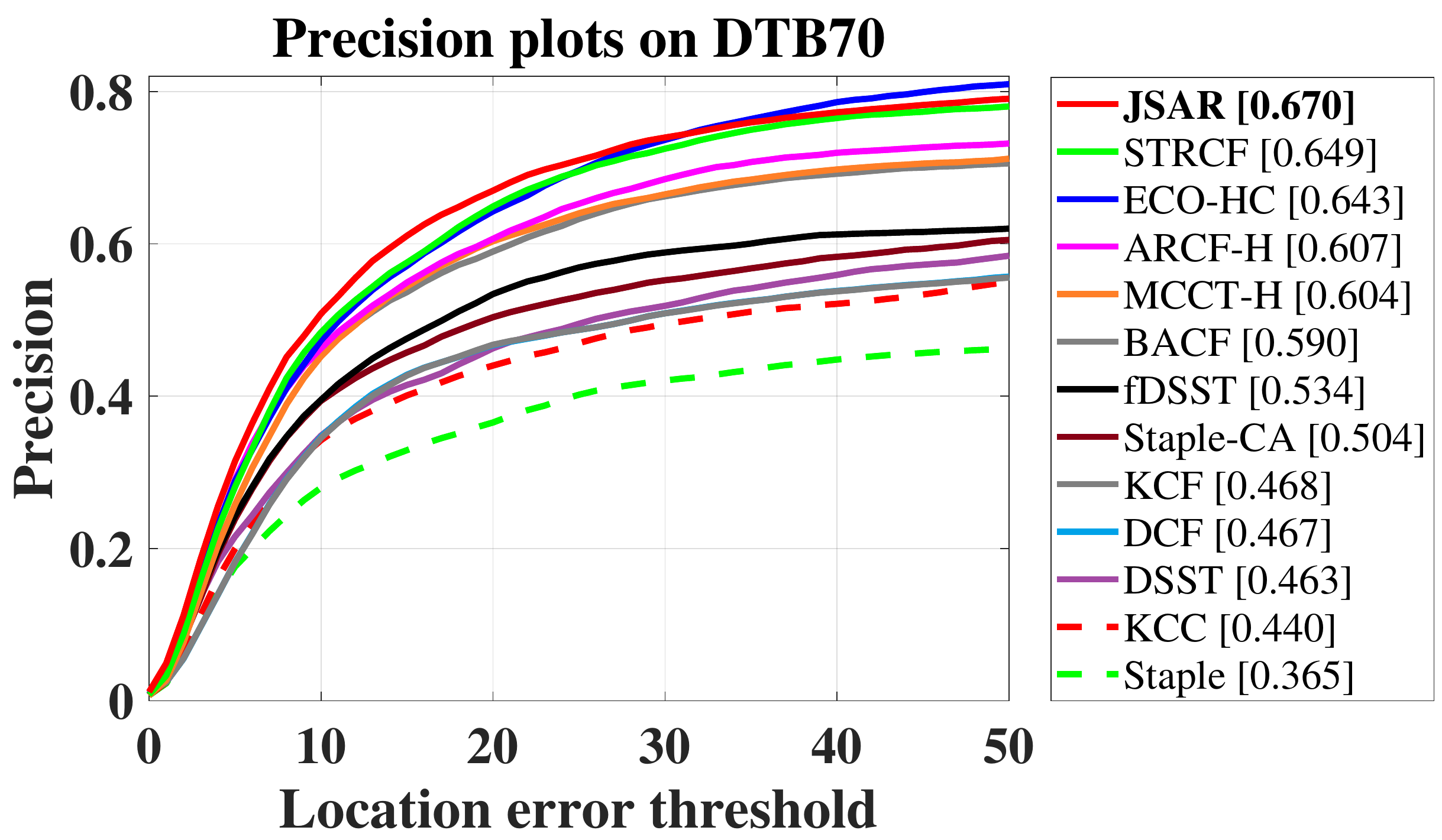}
				\\
				\includegraphics[width=1\columnwidth]{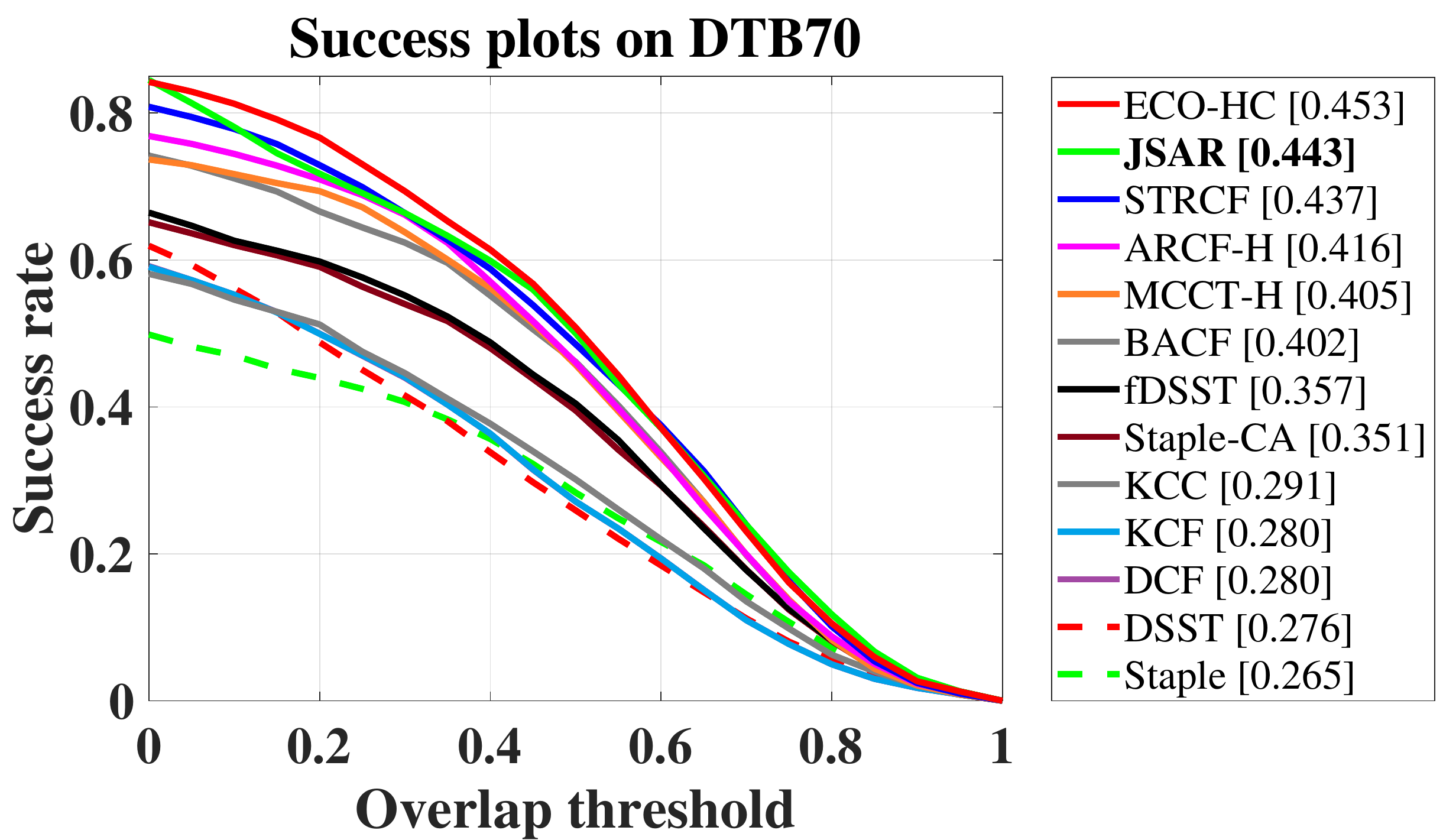}
			\end{minipage}
		}
	\end{center}
	\caption{Overall performance of hand-crafted real-time trackers on  (a) UAVDT~\cite{du2018unmanned} (b) UAV123@10fps~\cite{mueller2016benchmark} (c) DTB70~\cite{li2017visual}. JSAR has a notable improvement of 8.3\% and 4.3\% in terms of AUC on UAVDT and UAV123@10fps compared with the second best trackers, respectively. }
	\label{fig:overall}
\end{figure*}
\begin{table*}[t]
	\centering
	\setlength{\tabcolsep}{0.3mm}
	\fontsize{8}{9}\selectfont
	\begin{threeparttable}
		\caption{Average precision, AUC and speed comparison of top 10 hand-crafted trackers on UAVDT\cite{du2018unmanned}, UAV123@10fps\cite{mueller2016benchmark} and DTB70\cite{li2017visual}. \textcolor[rgb]{ 1,  0,  0}{Red}, 
			\textcolor[rgb]{ 0,  1,  0}{green} and \textcolor[rgb]{ 0,  0,  1}{blue} respectively mean the first, second and third place.}
		\vspace{0.08cm}
		\begin{tabular}{cccccccccccc}
			\toprule[2pt]
			\multirow{2}{*}{Tracker}&
			\multicolumn{6}{c}{Real time}&\multicolumn{4}{c}{Non-real time}\cr
			\cmidrule(lr){2-7} \cmidrule(lr){8-11}
			&\textbf{JSAR}&MCCT-H\cite{wang2018multi}&STRCF\cite{li2018CVPR}&ARCF-H\cite{huang2019learning}&BACF\cite{galoogahi2017learning}&ECO-HC\cite{danelljan2017eco}&CSR-DCF\cite{lukezic2017discriminative}&SRDCF\cite{danelljan2015learning}&ARCF-HC\cite{huang2019learning}&SRDCFdecon\cite{danelljan2016adaptive}\cr
			\midrule
			AUC&\textcolor[rgb]{ 0,  1,  0}{\textbf{0.465}}&0.413&0.435&0.421&0.416&\textcolor[rgb]{ 0,  0,  1}{\textbf{0.442}}&0.426&0.416&\textcolor[rgb]{1,  0,  0}{\textbf{0.468}}&0.397\cr
			Precision&\textcolor[rgb]{ 0,  1,  0}{\textbf{0.689}}&0.622&0.635&0.641&0.616&0.653&\textcolor[rgb]{ 0,  0,  1}{\textbf{0.654}}&0.616&\textcolor[rgb]{ 1,  0,  0}{\textbf{0.693}}&0.577\cr
			Speed(fps)&\textbf{32.2}&\textcolor[rgb]{ 0,  1,  0}{\textbf{59.7}}&28.5&51.2&\textcolor[rgb]{ 0,  0,  1}{\textbf{56.0}}&\textcolor[rgb]{1,  0,  0}{\textbf{69.3}}&12.1&14.0&15.3&7.5\cr
			Conference&\textbf{This work.}&CVPR'18&CVPR'18&ICCV'19&CVPR'17&CVPR'17&CVPR
			17&ICCV'15&ICCV'19&CVPR'16\cr
			\bottomrule[2pt]
			\label{tab:2}
		\end{tabular}
	\end{threeparttable}
\end{table*}
\begin{table}[t]
	\centering
	\setlength{\tabcolsep}{0.9mm}
	\fontsize{8}{10}\selectfont
	\begin{threeparttable}
		\caption{Precision, AUC and speed comparison between 14 recent deep trackers on UAVDT\cite{du2018unmanned}. \textcolor[rgb]{ 1,  0,  0}{Red}, 
			\textcolor[rgb]{ 0,  1,  0}{green}, \textcolor[rgb]{ 0,  0,  1}{blue} and  \textcolor[rgb]{ 1, 0.65 ,  0}{orange} respectively mean the first, second, third and fourth place.}
		\vspace{0.08cm}
		\begin{tabular}{cccccccccccc}
			\toprule[2pt]
			Tracker&AUC&Precision&Speed(fps)&CPU/GPU&Conference\\
			\midrule
			\textbf{JSAR}&\textcolor[rgb]{1,0,0}{\textbf{0.469}}&\textcolor[rgb]{1,0,0}{\textbf{0.721}}&\textcolor[rgb]{1,0.65,0}{\textbf{35}}&\textbf{CPU}&\textbf{This work.}\\
			GOTURN\cite{held2016learning}&0.451&\textcolor[rgb]{1,0.65,0}{\textbf{0.702}}&17&GPU&ECCV'16\\
			IBCCF\cite{li2017integrating}&0.388&0.603&3&GPU&CVPR'17\\
			TADT\cite{li2019target}&0.431&0.677&\textcolor[rgb]{1,0.65,0}{\textbf{35}}&GPU&CVPR'19\\
			DSiam\cite{guo2017learning}&\textcolor[rgb]{0,0,1}{\textbf{0.457}}&\textcolor[rgb]{0,0,1}{\textbf{0.704}}&16&GPU&ICCV'17\\
			PTAV\cite{fan2017parallel}&0.384&0.675&27&GPU&ICCV'17\\
			ECO\cite{danelljan2017eco}&\textcolor[rgb]{1,0.65,0}{\textbf{0.454}}&0.700&16&GPU&CVPR'17\\
			ASRCF\cite{dai2019visual}&0.437&0.700&24&GPU&CVPR'19\\
			MCCT\cite{wang2018multi}&0.437&0.671&9&GPU&CVPR'18\\
			CFNet\cite{valmadre2017CVPR}&0.428&0.680&\textcolor[rgb]{0,1,0}{\textbf{41}}&GPU&CVPR'17\\
			C-COT\cite{danelljan2016beyond}&0.406&0.656&1&GPU&ECCV'16\\
			ADNet\cite{yun2017action}&0.429&0.683&8&GPU&CVPR'17\\
			UDT+\cite{wang2019unsupervised}&0.416&0.697&\textcolor[rgb]{1,0,0}{\textbf{60}}&GPU&CVPR'19\\
			SiameseFC\cite{bertinetto2016fully}&\textcolor[rgb]{0,1,0}{\textbf{0.465}}&\textcolor[rgb]{0,1,0}{\textbf{0.708}}&\textcolor[rgb]{0,0,1}{\textbf{38}}&GPU&ECCV'16\\
			DeepSTRCF\cite{li2018CVPR}&0.437&0.667&6&GPU&CVPR'18\\
			\bottomrule[2pt]
		\end{tabular}
		\label{tab:1}
	\end{threeparttable}
\end{table}
In this section, the proposed method is evaluated on three challenging short-term UAV benchmarks, \emph{i.e.}, UAVDT\cite{du2018unmanned}, UAV123@10fps\cite{mueller2016benchmark}, DTB70\cite{li2017visual} and one long-term benchmark UAV20L\cite{mueller2016benchmark}, including over 149K images overall captured by drone camera in all kinds of harsh aerial scenarios. The experimental result of our method is compared with 30 state-of-the-art (SOTA) approaches including 14 deep trackers, \emph{i.e.}, SiameseFC\cite{bertinetto2016fully}, DSiam\cite{guo2017learning}, IBCCF\cite{li2017integrating}, ECO\cite{danelljan2017eco}, C-COT\cite{danelljan2016beyond}, GOTURN\cite{held2016learning}, PTAV\cite{fan2017parallel}, DeepSTRCF\cite{li2017integrating}, 	CFNet\cite{valmadre2017CVPR}, ASRCF\cite{dai2019visual}, MCCT\cite{wang2018multi}, ADNet\cite{yun2017action}, TADT\cite{li2019target}, UDT+\cite{wang2019unsupervised}, and 16 hand-crafted trackers, \emph{i.e.}, MCCT-H\cite{wang2018multi}, KCF\cite{henriques2015high}, DSST\cite{danelljan2014accurate}, fDSST\cite{danelljan2017discriminative}, ECO-HC\cite{danelljan2017eco}, DCF\cite{henriques2015high}, BACF\cite{galoogahi2017learning}, ARCF\cite{huang2019learning}, SRDCF\cite{danelljan2015learning}, STAPLE-CA\cite{mueller2017context}, ARCF-H\cite{huang2019learning}, STAPLE\cite{bertinetto2016staple}, SRDCFdecon\cite{danelljan2016adaptive}, CSR-DCF\cite{lukezic2017discriminative}, KCC\cite{wang2018kernel}, STRCF\cite{li2018CVPR}.
\subsection{Implementation details}
To test the size estimation ability of JSAR, first of all, experiments are conducted on three short-term UAV benchmarks\cite{mueller2016benchmark,li2017visual,du2018unmanned}, compared with both deep and hand-crafted trackers. Then, the re-detection module is added to cope with tracking failure, generating JSAR-Re. We evaluate JSAR-Re with SOTA trackers on UAV20L dataset\cite{mueller2016benchmark}.
\subsubsection{\textbf{Platform}}
All experiments are implemented with MATLAB R2018a and all experimental results are obtained on a computer with a single i7-8700K (3.70GHz) CPU, 32GB RAM,
and an NVIDIA RTX 2080 GPU for fair comparisons.
\subsubsection{\textbf{Baseline}}
In this work, spatial-temporal regularized correlation filter (STRCF) \cite{li2018CVPR} is selected as our baseline tracker which adds spatio-temporal regularized term to training objective for improving robustness and adopts a multi-scale search strategy for scale adaptivity. Discarding the hierarchical scale searching, JSAR separately trains a size filter using Eq.~(\ref{eqn:2}) to estimate the scale and aspect ratio variations and follows the translation estimation in \cite{li2018CVPR}.
\subsubsection{\textbf{Features}}
To guarantee real-time performance on a low-cost CPU, we only apply hand-crafted features to our tracker for experiments. Gray-scale, histogram of oriented gradient (HOG) \cite{henriques2015high} and color name (CN) \cite{Danelljan2014CVPR} are employed in translation filter, while size filter only uses HOG.
\subsubsection{\textbf{Hyper Parameters}}
The main parameters in this work are listed in Table \ref{tab:4}. For impartial comparison, all the parameters are fixed in the experiments. 
\renewcommand{\arraystretch}{1.5} 
\subsubsection{\textbf{Criteria}}
Following one-pass evaluation (OPE) \cite{wu2013online} protocol, we evaluate all trackers by two measures, \emph{i.e.}, precision and success rate. Precision plots can exhibit the percentage of all input images in which the distance of predicted location with ground truth one is smaller than various thresholds, and success
plots can reflect the proportion of frames in which the intersection over union (IoU) between the estimated bounding box and the ideal one is greater than distinctive thresholds. The score at 20 pixel and area under curve (AUC) are respectively used to rank the trackers. 
\subsection{JSAR vs. deep trackers}
\begin{figure*}[!t]
	\centering	
	\includegraphics[width=0.97\textwidth]{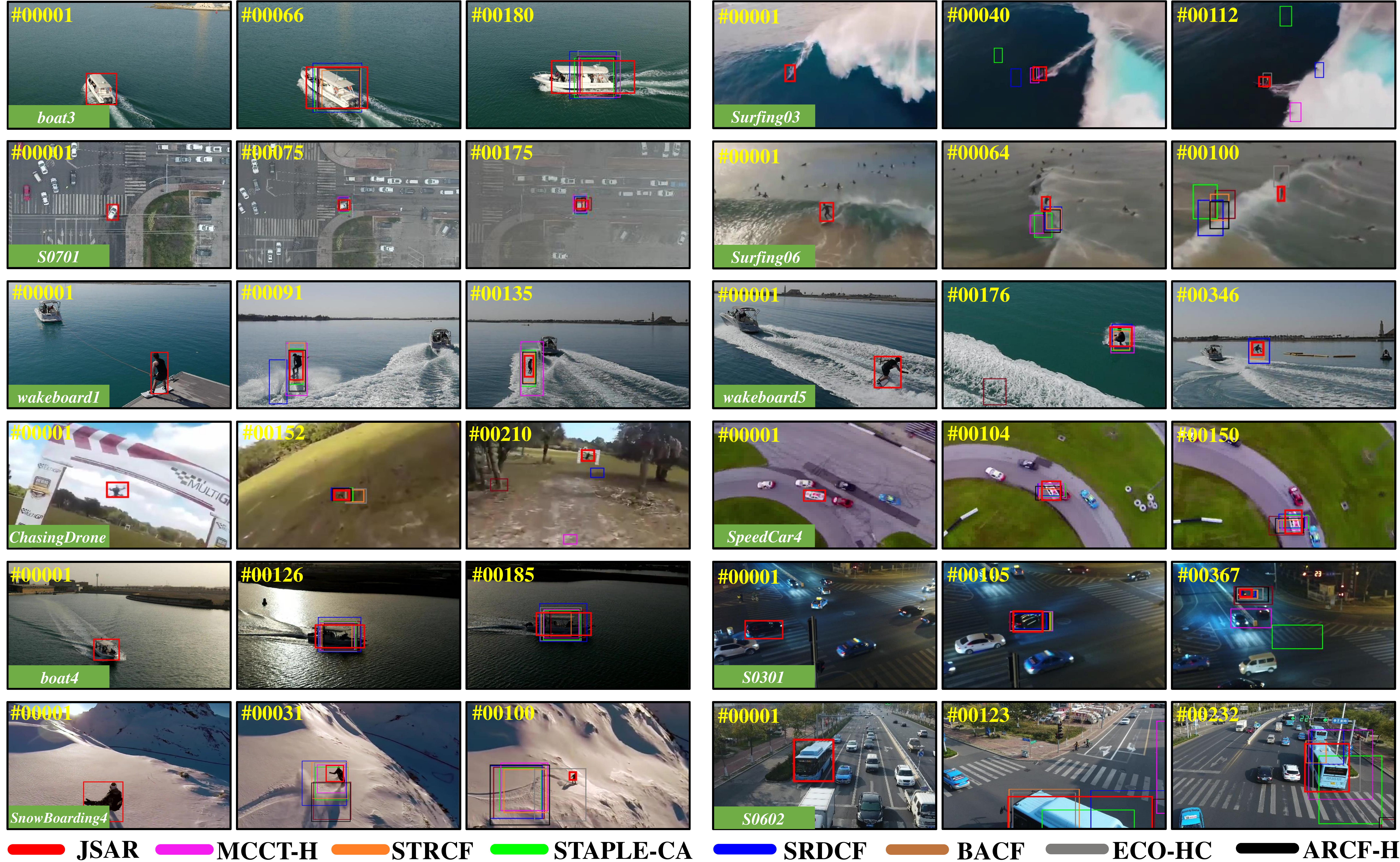}
	\caption{Display of tracking results from eight hand-crafted trackers on twelve UAV video, \emph{i.e.}, \emph{S0301}, \emph{S0602}, \emph{S0701} of UAVDT~\cite{du2018unmanned}, \emph{boat3}, \emph{boat4}, \emph{wakeboard1}, \emph{wakeboard5} of UAV123@10fps~\cite{mueller2016benchmark} and \emph{ChasingDrones}, \emph{SnowBoarding4}, \emph{Surfing03} \emph{Surfing06},  \emph{SpeedCar4} of DTB70~\cite{li2017visual}. }
	\label{fig:display}
\end{figure*}

We first compare the tracking performance of JSAR with 14 recently proposed SOTA deep trackers, \emph{i.e.}, deep features based trackers and deep convolution neural networks (DCNN) based trackers, on UAVDT benchmark\cite{du2018unmanned}. As shown in Table~\ref{tab:1}, JSAR has taken the first place in both precision and AUC, while coming fourth in tracking speed running on a low-cost CPU. Without robust deep features, the remarkable improvement (7.3\% and 8.1\% than DeepSTRCF in terms of AUC and precision) can be attributed to the ARC adaption, because UAVDT mainly addresses vehicle tracking and the viewpoint change can easily lead to ARC of the tracked vehicle in the image, as shown in Figure \ref{fig:display}.
\subsection{JSAR vs. hand-crafted trackers}
\subsubsection{Overall evaluation}
\begin{figure*}[!t]
	\begin{center}

		\subfigure { \label{fig:att1} 
			\begin{minipage}{0.23\textwidth}
				\centering
				\includegraphics[width=1\columnwidth]{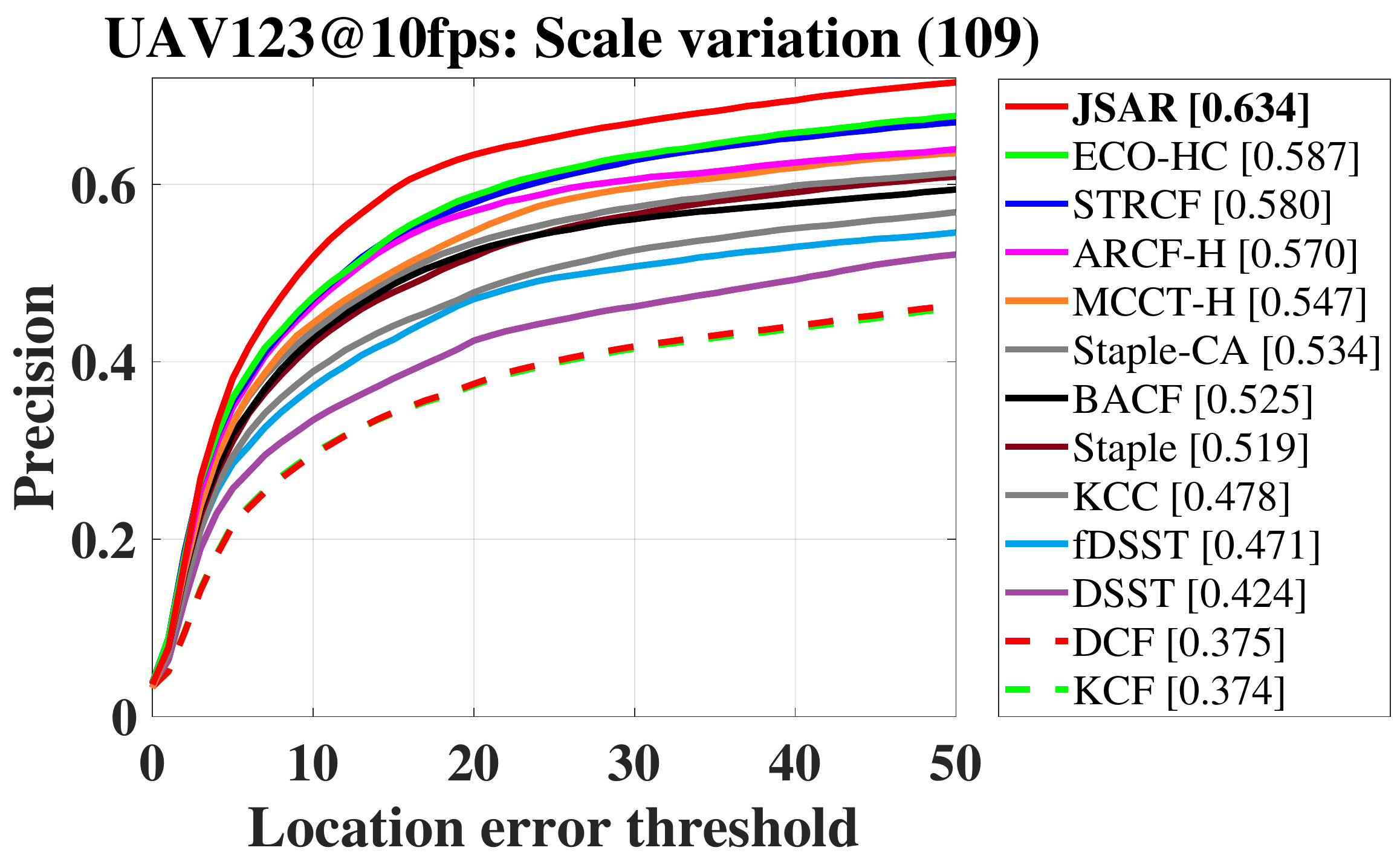}
			\end{minipage}
		}
		\subfigure { \label{fig:att2} 
			\begin{minipage}{0.23\textwidth}
				\centering
				\includegraphics[width=1\columnwidth]{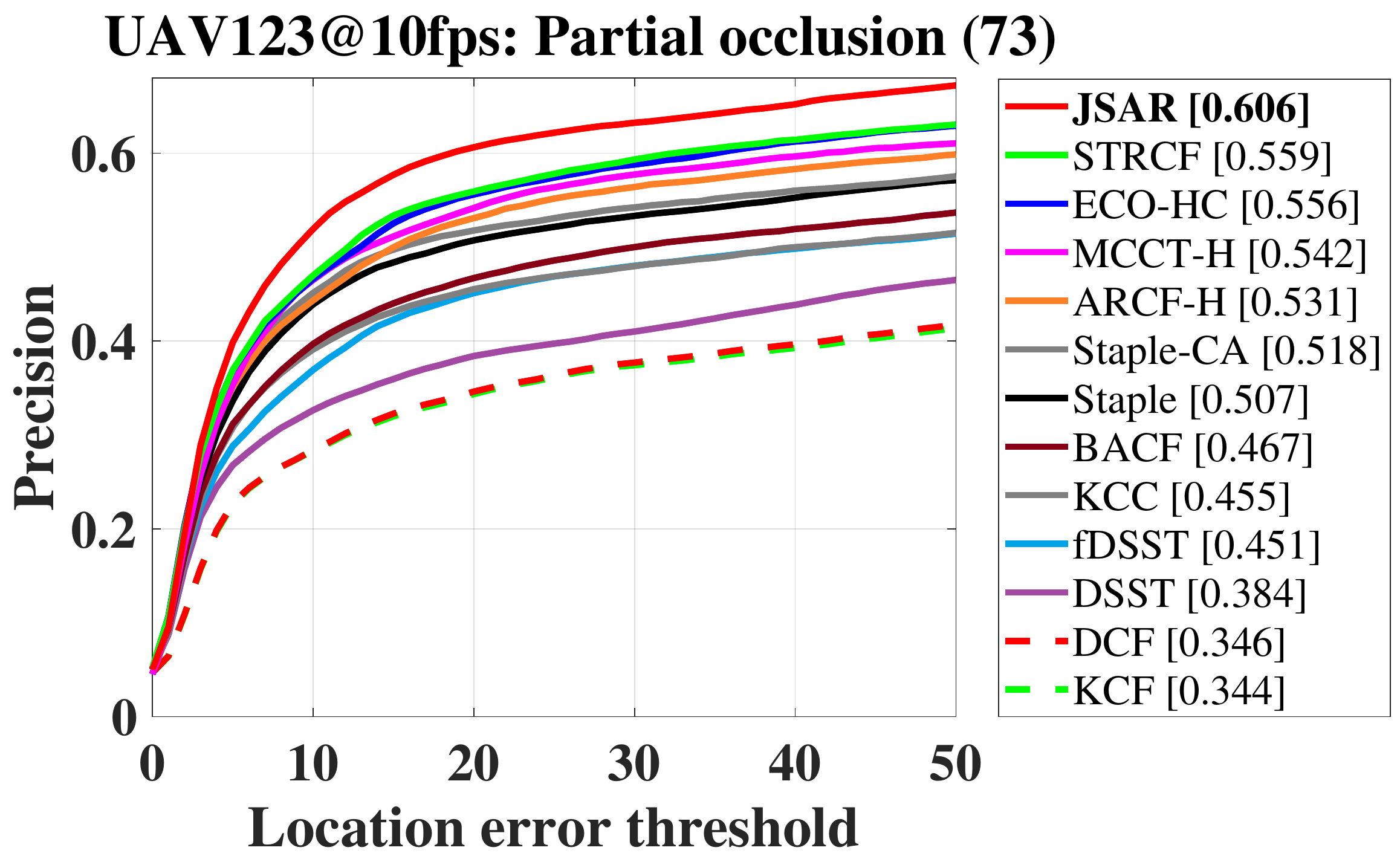}
			\end{minipage}
		}
		\subfigure { \label{fig:att3} 
			\begin{minipage}{0.23\textwidth}
				\centering
				\includegraphics[width=1\columnwidth]{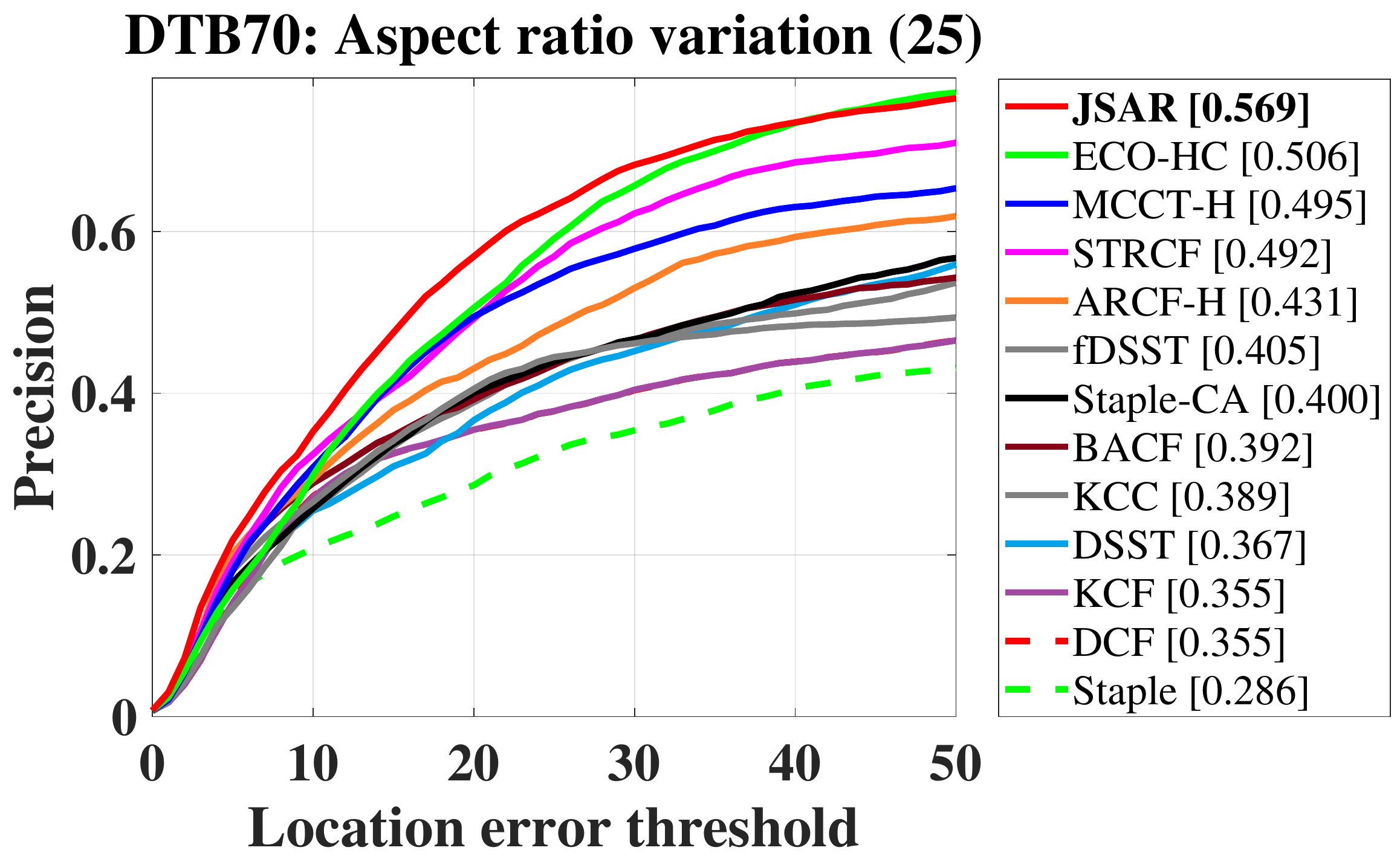}
			\end{minipage}
		}
		\subfigure { \label{fig:att4} 
			\begin{minipage}{0.23\textwidth}
				\centering
				\includegraphics[width=1\columnwidth]{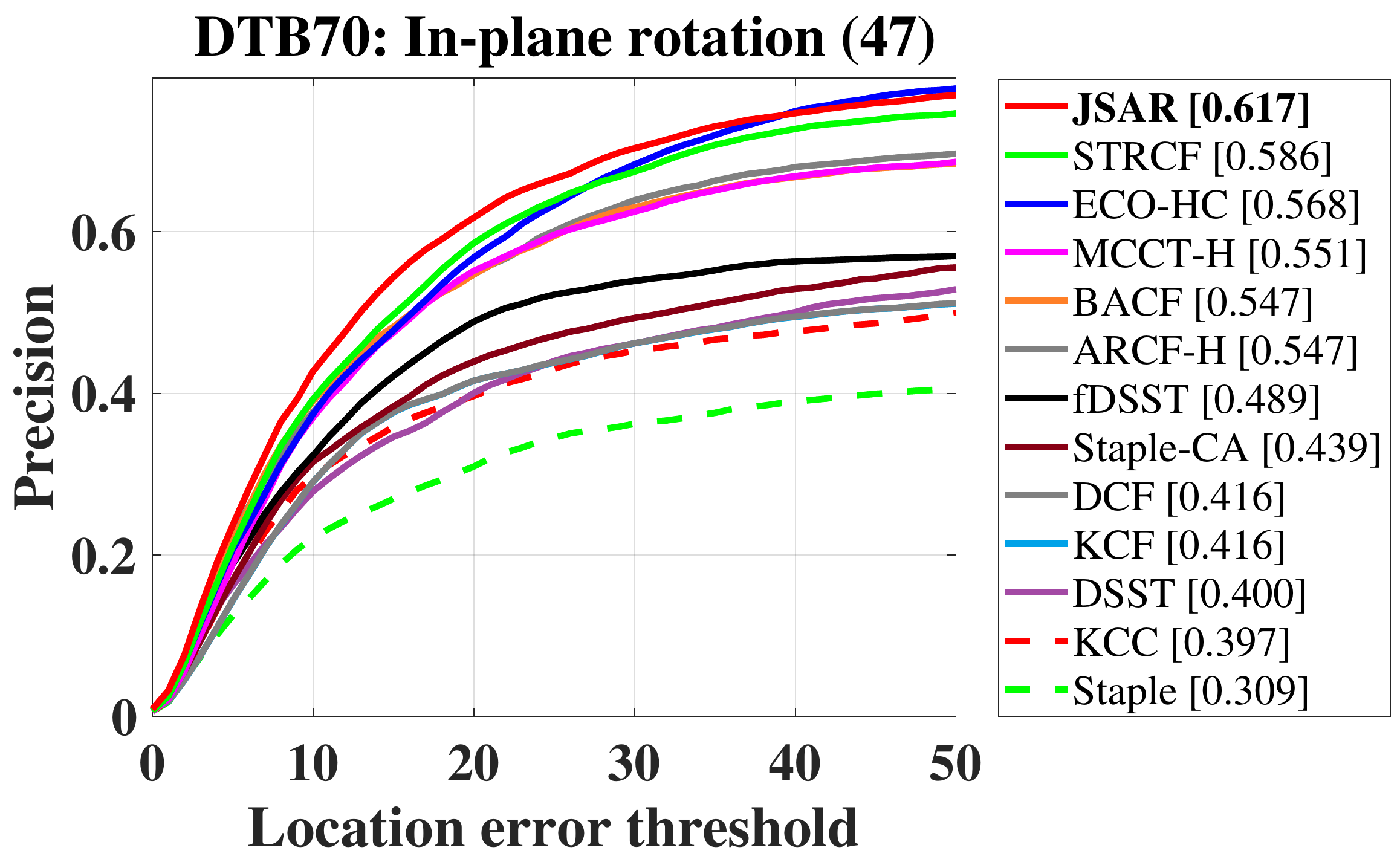}
			\end{minipage}
		}
		\subfigure { \label{fig:att5} 
			\begin{minipage}{0.23\textwidth}
				\centering
				\includegraphics[width=1\columnwidth]{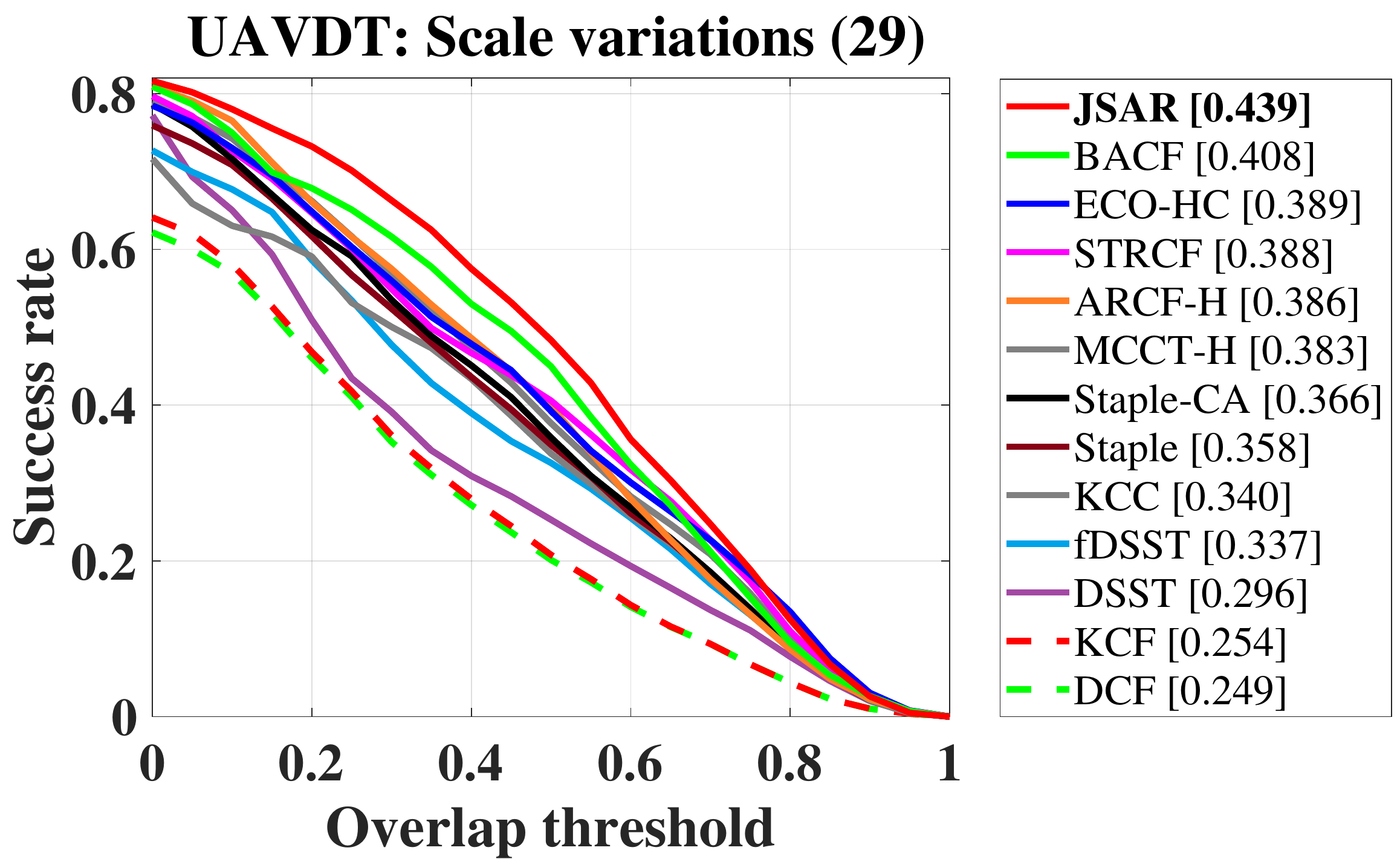}
			\end{minipage}
		}
		\subfigure { \label{fig:att6} 
			\begin{minipage}{0.23\textwidth}
				\centering
				\includegraphics[width=1\columnwidth]{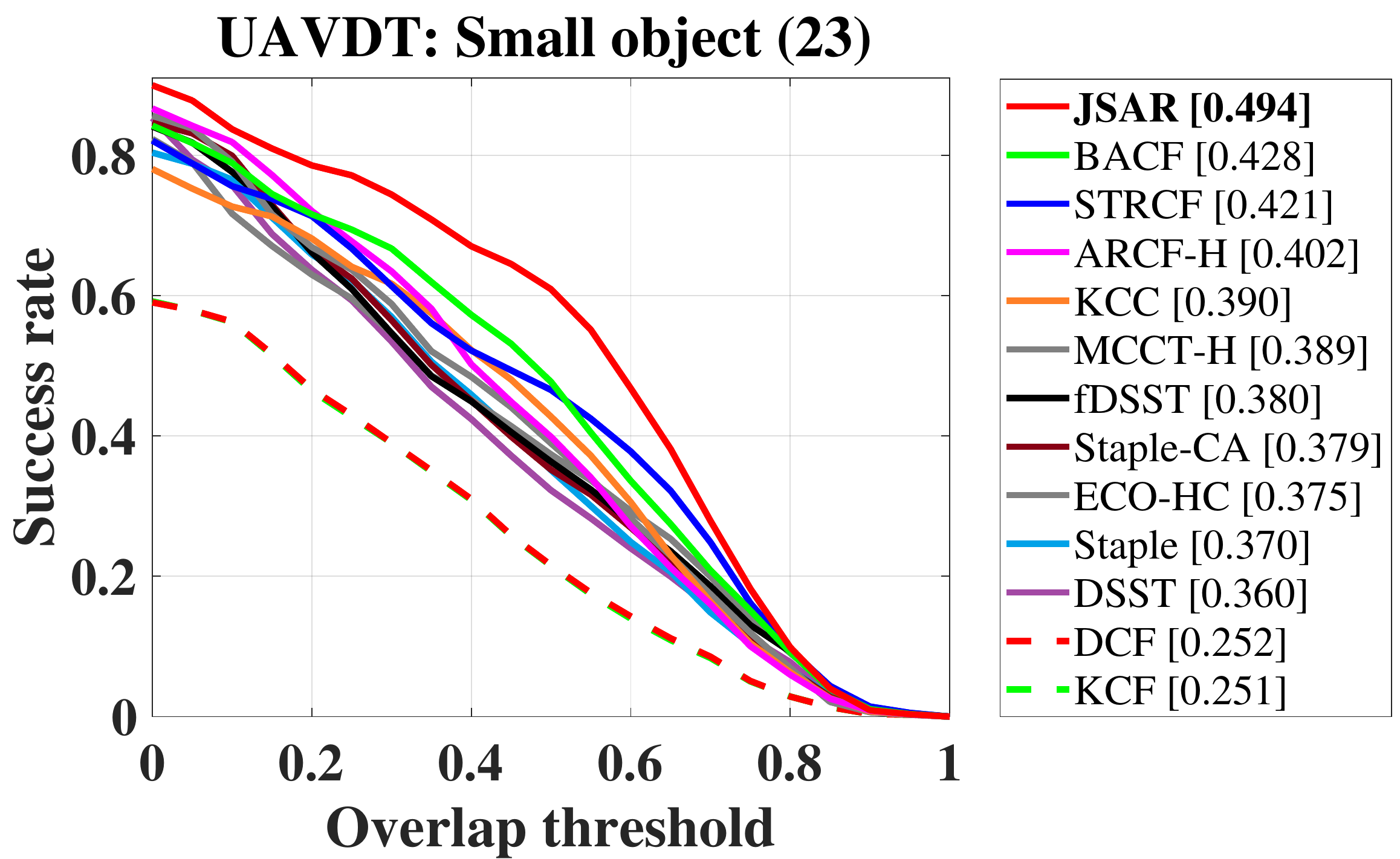}
			\end{minipage}
		}
		\subfigure { \label{fig:att7} 
			\begin{minipage}{0.23\textwidth}
				\centering
				\includegraphics[width=1\columnwidth]{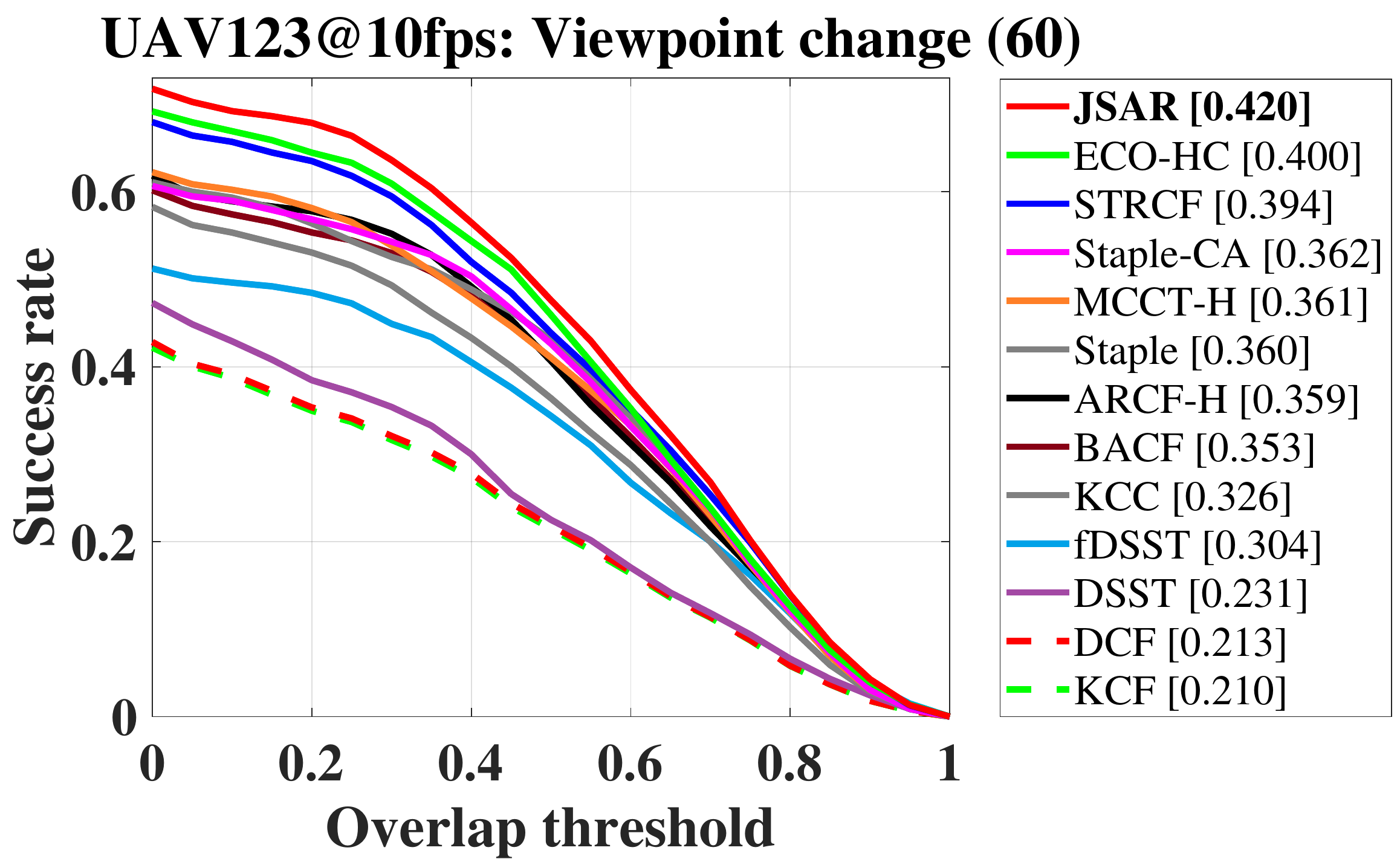}
			\end{minipage}
		}
		\subfigure { \label{fig:att8} 
			\begin{minipage}{0.23\textwidth}
				\centering
				\includegraphics[width=1\columnwidth]{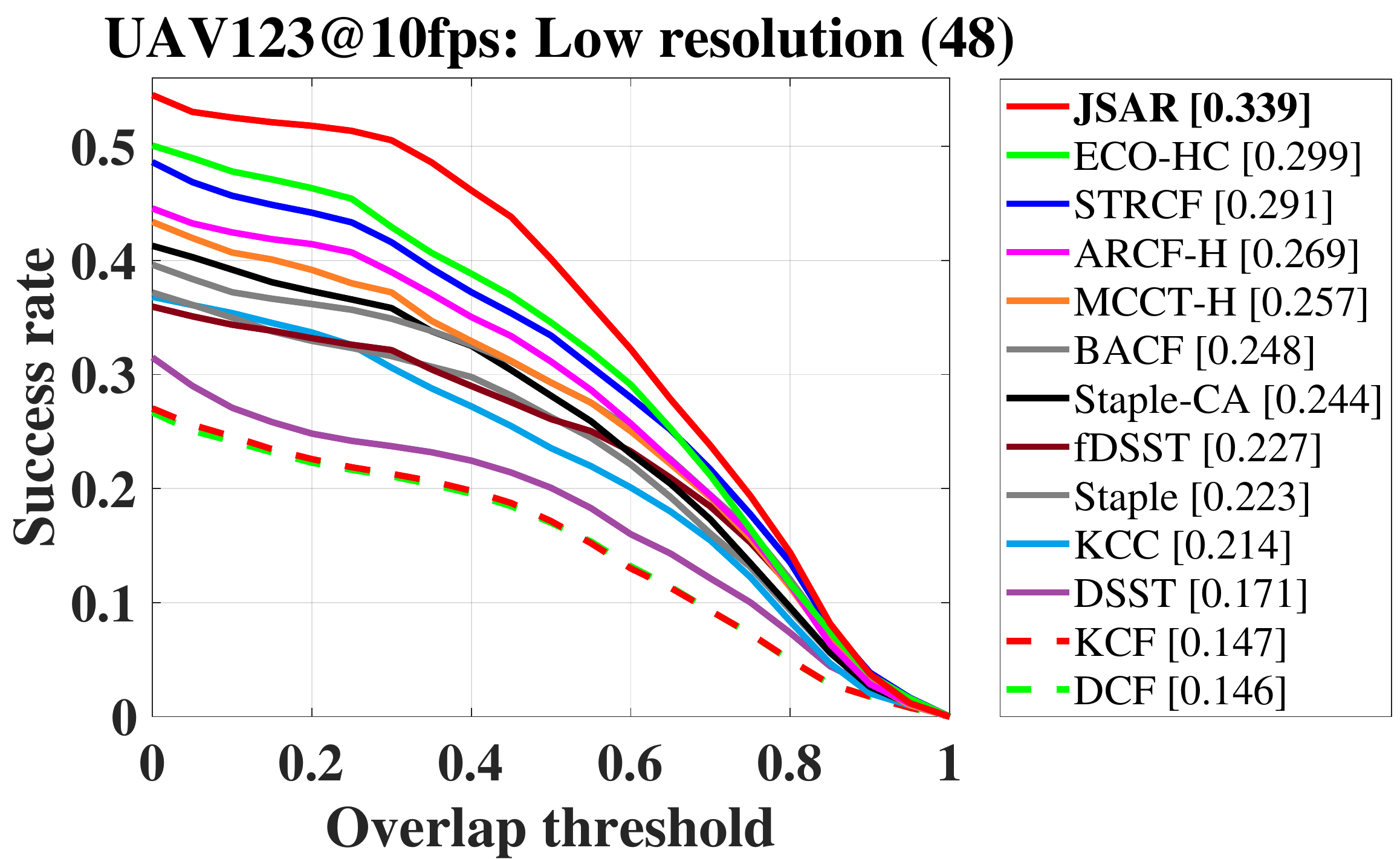}
			\end{minipage}
		}
	\end{center}
	
	\caption{Attribute-oriented comparison with hand-crafted real-time  trackers. Precision plots of four attributes, \emph{i.e.}, scale variation, partial occlusion, aspect ratio variation, in-plane rotation, and success plots of four attribute, \emph{i.e.}, scale variations, small object, viewpoint change and low resolution are presented.}
	\label{fig:attribute}
\end{figure*}
Restricted by scarce computation resources, deep trackers have difficulties meeting real-time tracking speed on UAV. Hand-crafted trackers, \emph{i.e.}, using hand-crafted features in DCF framework, are ideal choices in UAV tracking for its calculation efficiency. In this subsection, twelve SOTA and real-time hand-crafted trackers are used for comparison with JSAR at first. JSAR outperforms other real-time trackers in terms of precision and AUC, as displayed in Fig.~\ref{fig:overall}. Notably, compared with the baseline STRCF\cite{li2018CVPR}, JSAR respectively improves the AUC by 14.1\%, 5.5\%, and the precision by 14.6\%, 7.7\%  on UAVDT and UAV123@10fps. We further compare the average performance of best 10 hand-crafted trackers on three benchmarks\cite{mueller2016benchmark,du2018unmanned,li2017visual}, as shown in Table~\ref{tab:2}. It can be seen JSAR obtained the second place in both AUC and precision, however, JSAR has a tiny gap compared with the best tracker ARCF-HC\cite{huang2019learning} (0.003 and 0.004 in AUC and precision), and it has remarkably improved the speed by 110\%. Hence, compared to ARCF-HC, JSAR performs comparably with much higher efficiency. Averagely, JSAR gains an improvement of 6.9\% in AUC and 8.5\% in precision compared with the baseline method, \emph{i.e.}, STRCF.
\subsubsection{Attribute-oriented evaluation}
Fig. \ref{fig:attribute} exhibits the 
precision and success plots of real-time trackers on eight challenging attributes from UAV123@10fps\cite{mueller2016benchmark}, UAVDT\cite{du2018unmanned} and DTB70\cite{li2017visual}. It can be seen that JSAR has respectively improved the precision by 8\%, 12.5\%, and 5.3\% compared with the second-best trackers in the attributes of scale variation, aspect ratio variation, and in-plane rotation. As for AUC, JSAR gains a 7.6\% improvement in scale variations and a 5.0\% improvement in viewpoint change. The remarkable improvements demonstrate the effectiveness of the size filter in scale and aspect ratio change cases. Besides, in partial occlusion, small object, and low resolution, JSAR still outperforms other real-time trackers dramatically, exhibiting its excellent generality in various aerial scenarios.
\subsection{Hyper parameters analysis}
\begin{figure}[!b]
	\begin{center}

		\subfigure { \label{fig:hyper1} 
			\begin{minipage}{0.48\textwidth}
				\centering
				\includegraphics[width=1\columnwidth]{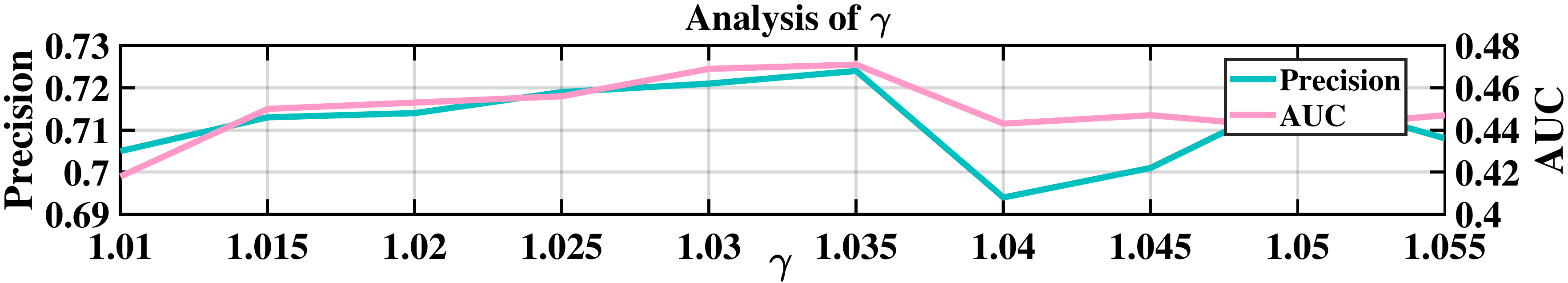}
			\end{minipage}
		}
		\subfigure { \label{fig:hyper2} 
			\begin{minipage}{0.48\textwidth}
				\centering
				\includegraphics[width=1\columnwidth]{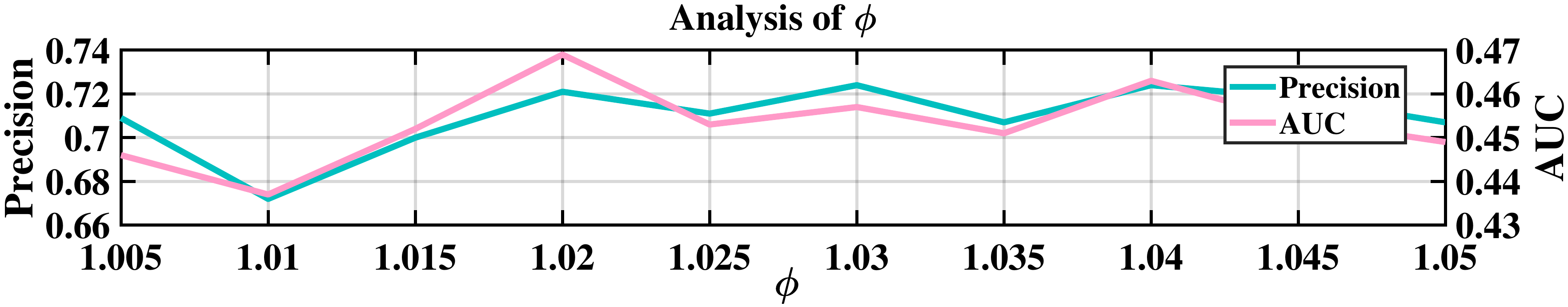}
			\end{minipage}
		}
		\subfigure { \label{fig:hyper3} 
			\begin{minipage}{0.48\textwidth}
				\centering
				\includegraphics[width=1\columnwidth]{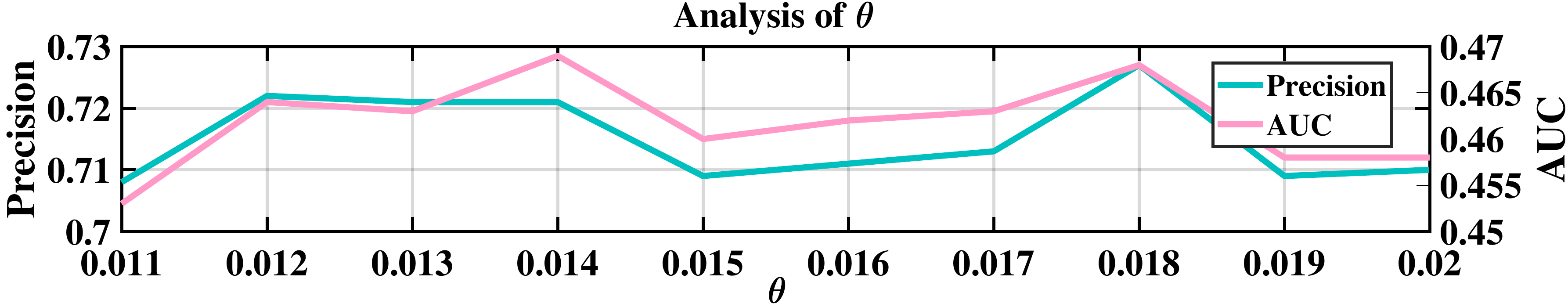}
			\end{minipage}
		}
	\end{center}
	\caption{Sensitivity analysis of four parameters ($\gamma$, $\phi$,
		,$\theta$ and $S/A$) on UAVDT\cite{du2018unmanned}. It is noted that we fixed the untested parameters in analysis.}
	\label{fig:hyper}
\end{figure}

We analyze the impacts of five core hyper parameters in the proposed size filter, including the sampling step $\gamma$ and $\phi$, the learning rate of size filter $\theta$, and the number of scale as well as aspect ratio (S/A). The impacts on AUC and precision of the first three parameters are exhibited in Figure~\ref{fig:hyper}, from which it can be seen they have a relatively small influence on tracking performance (with precision from 0.672 to 0.721, AUC from 0.418 to 0.469), which demonstrates the strong robustness of JSAR. The comparison of tracking performance and speed of various S/A configurations are displayed in Table~\ref{tab:3}. From 9 to 21, the number of scales/aspect ratios has little influence on both AUC (ranging from 0.452 to 0.469) and precision (ranging from 0.704 to 0.721). Yet the results rapidly fall off when the value of S/A is 7. This situation can be explained by insufficient samples for size filter training. 
\begin{table}[!b]
	\centering
	\setlength{\tabcolsep}{0.8mm}
	\fontsize{8}{6}\selectfont
	\begin{threeparttable}
		\caption{Demographic Prediction performance comparison by three evaluation metrics.}
		\vspace{0.08cm}
		\begin{tabular}{cccccccccc}
			\toprule[2pt]
		    S/A&5&7&9&11&13&15&17&19&21\\
			\midrule
		     AUC&0.373&0.397&0.452&0.465&\bf{0.469}&0.454&0.455&0.452&0.454\\
		     Precison&0.696&0.685&0.715&0.719&\bf{0.721}&0.704&0.712&0.709&0.713\\
		     Speed(fps)&50.2&46.9&43.1&38.5&\bf{35.1}&31.1&27.7&24.0&21.7\\
			\bottomrule[2pt]
			\label{tab:3}
		\end{tabular}
	\end{threeparttable}
\end{table}
\subsection{Re-detection evaluation}
\begin{figure}[!t]
	\begin{center}
		\centering
		\includegraphics[width=0.93\columnwidth]{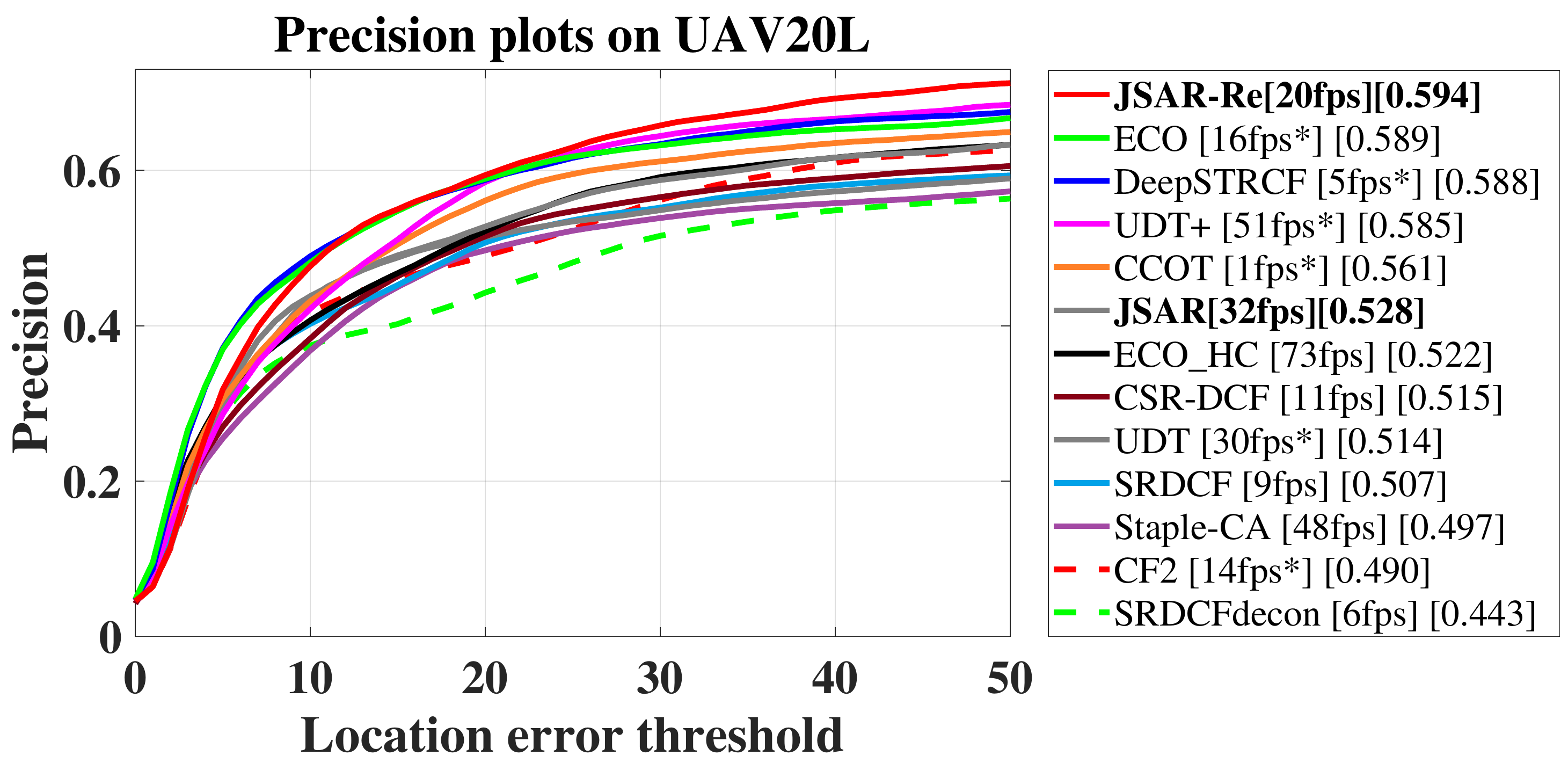}
	\end{center}
	\caption{Precision plots with tracking speed of JSAR-Re, JSAR and eleven SOTA trackers on UAV20L dataset \cite{mueller2016benchmark}. * denotes this tracker is tested on GPU.}
	\label{fig:UAV20L}
\end{figure}

\begin{figure}[!t]
	\begin{center}
		\centering
		\includegraphics[width=1\columnwidth]{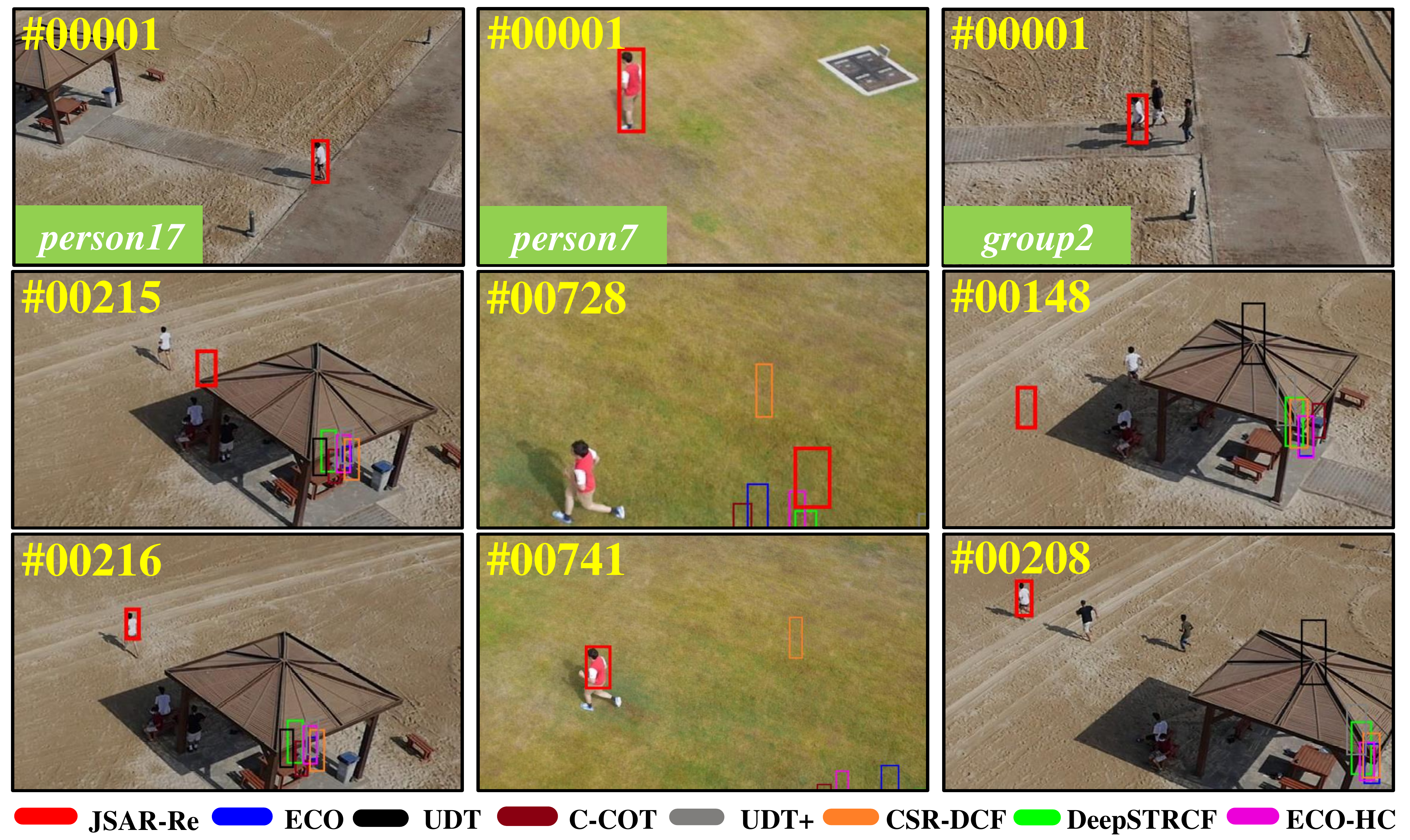}
	\end{center}
	\caption{Qualitative tracking performance of JSAR and seven SOTA trackers on \emph{person14}, \emph{person7} and \emph{group2} of UAV20L dataset\cite{mueller2016benchmark}.}
	\label{fig:long-term}
\end{figure}

To testify the effectiveness of our proposed re-detection strategy, we conduct experiments on JSAR-Re and JSAR with eleven SOTA trackers on long-term UAV20L benchmarks, which consists of 20 long-term image sequences with over 2.9K frames per sequence averagely. The precision plot is reported in Figure~\ref{fig:UAV20L}. JSAR-Re ranks No.1 and improves the tracking precision by 11.3\% compared with JSAR, with a speed of 20fps on a low-cost CPU. Some qualitative results are exhibited in Figure~\ref{fig:long-term}.
\section{CONCLUSIONS}\label{sec:CONCLUSIONS}
In this work, a novel UAV tracking framework of joint scale and ARC estimation is proposed. Also, an object proposal based re-detection algorithm is introduced to achieve long-term tracking. Experimental comparison with 30 SOTA trackers exhibits the superiority of our method. Most tellingly, our method can outperform SOTA deep trackers on UAVDT\cite{du2018unmanned} with only hand-crafted features. Using C++ implementation can further raise the tracking speed for real-world as well as real-time UAV applications.


\section*{ACKNOWLEDGMENT}
This work is supported by the National Natural Science Foundation of China (No. 61806148).


\bibliographystyle{IEEEtran}  
\bibliography{IEEEabrv,ref}

\end{document}